\documentclass[a4paper,fleqn]{cas-sc}

\usepackage[authoryear,longnamesfirst]{natbib}
\usepackage{times}
\usepackage{latexsym}
\usepackage{multirow}
\usepackage{graphicx}
\usepackage{enumitem}
\usepackage[linesnumbered,ruled,vlined]{algorithm2e}
\usepackage{amsmath}
\usepackage{latexsym}
\usepackage[inkscapelatex=false]{svg}
\usepackage{url}

\usepackage{ulem}   
\usepackage{subfigure}
\usepackage{makecell}
\usepackage{color}
\usepackage{booktabs}
\usepackage{hyperref}
\usepackage[all]{nowidow}   
\usepackage[subtle]{savetrees} 

\SetKwInput{KwInput}{Input}                
\SetKwInput{KwOutput}{Output}              

\def\tsc#1{\csdef{#1}{\textsc{\lowercase{#1}}\xspace}}
\tsc{WGM}
\tsc{QE}
\tsc{EP}
\tsc{PMS}
\tsc{BEC}
\tsc{DE}

\begin{document}
\let\WriteBookmarks\relax
\def\floatpagepagefraction{1}
\def\textpagefraction{.001}

\shorttitle{A Cross-Attention Augmented Model for Event-Triggered Context-Aware Story Generation}

\shortauthors{Chen Tang et~al.}

\title [mode = title]{A Cross-Attention Augmented Model for Event-Triggered Context-Aware Story Generation}




\author[1]{Chen Tang}[style=Chinese,
    orcid=https://orcid.org/0000-0003-1188-4393]
\ead{chen.tang@surrey.ac.uk}

\author[2]{Tyler Loakman}[]
\ead{tcloakman1@sheffield.ac.uk}

\author[3]{Chenghua Lin}
\ead{chenghua.lin@manchester.ac.uk}
\cormark[1]

\affiliation[1]{organization={Department of Computer Science, University of Surrey},
    addressline={Stag Hill, University Campus, Guildford}, 
    city={Guildford},
    postcode={GU2 7XH}, 
    country={United Kingdom}}
    
\affiliation[2]{organization={Department of Computer Science, The University of Sheffield},
    addressline={Western Bank}, 
    city={Sheffield},
    postcode={S10 2TN}, 
    country={United Kingdom}}

\affiliation[3]{organization={Department of Computer Science, The University of Manchester},
    addressline={Oxford Road, Manchester, M13 9PL, United Kingdom}, 
    city={Manchester},
    postcode={S10 2TN}, 
    country={United Kingdom}}



\cortext[cor1]{Corresponding author}

\begin{abstract}
Despite recent advancements, existing story generation systems continue to encounter difficulties in effectively incorporating contextual and event features, which greatly influence the quality of generated narratives. To tackle these challenges, we introduce a novel neural generation model, EtriCA, that enhances the relevance and coherence of generated stories by employing a cross-attention mechanism to map context features onto event sequences through residual mapping. This feature capturing mechanism enables our model to exploit logical relationships between events more effectively during the story generation process. To further enhance our proposed model, we employ a post-training framework for knowledge enhancement (KeEtriCA) on a large-scale book corpus. This allows EtriCA to adapt to a wider range of data samples. This results in approximately 5\% improvement in automatic metrics and over 10\% improvement in human evaluation.
We conduct extensive experiments, including comparisons with state-of-the-art (SOTA) baseline models, to evaluate the performance of our framework on story generation. The experimental results, encompassing both automated metrics and human assessments, demonstrate the superiority of our model over existing state-of-the-art baselines. These results underscore the effectiveness of our model in leveraging context and event features to improve the quality of generated narratives.
\end{abstract}

\begin{highlights}
\item We introduce a novel task in the domain of event-driven story generation. This task necessitates the generation model to compose narratives based on a specified initial context and a sequence of events. 
\item We present an innovative method aimed at enhancing the existing event extraction framework. This enhancement is achieved by incorporating dependency parsing techniques. Furthermore, we provide annotated event sequences for two well-established datasets commonly used in our new task. 
\item We propose a neural generation model, \textbf{KeEtriCa}, which leverages the context and event sequence information with an enhanced cross-attention based feature capturing mechanism and sentence-level representation learning.
\item We conduct a series of experiments and a comprehensive analysis to investigate the underlying characteristics contributing to writing a more fluent, relevant, and coherent story.
\end{highlights}

\begin{keywords}
long text generation \sep story generation \sep event plan \sep context aware
\end{keywords}

\maketitle

\section{Introduction}
With the fast development of deep learning, artificial intelligence (AI) is able to handle increasingly challenging natural language generation tasks. For instance, the abilities of text generation models have increased from short texts (e.g. dialogues) to long texts (e.g. stories). An increasing number of studies~\citep{tang2022recent} have focused on improving the performance of neural network models for long text generation. The study of Story Generation aims to produce narratives that are fluent, relevant, and coherent, while being conditioned on a provided context. As the task is notoriously difficult, a prevalent approach involves utilising storylines comprising events to facilitate the generation process~\citep{chen-etal-2021-graphplan, alhussain2021automatic}. This strategy emulates the creative process observed in human writers~\citep{tang2022ngep}. Initially, a story commences with a skeletal outline consisting of essential keywords that represent key events. Subsequently, human writers gradually unfold the narrative by following the planned sequence of events.

\begin{figure}[ht]
\centering
\includegraphics[width=0.95\linewidth]{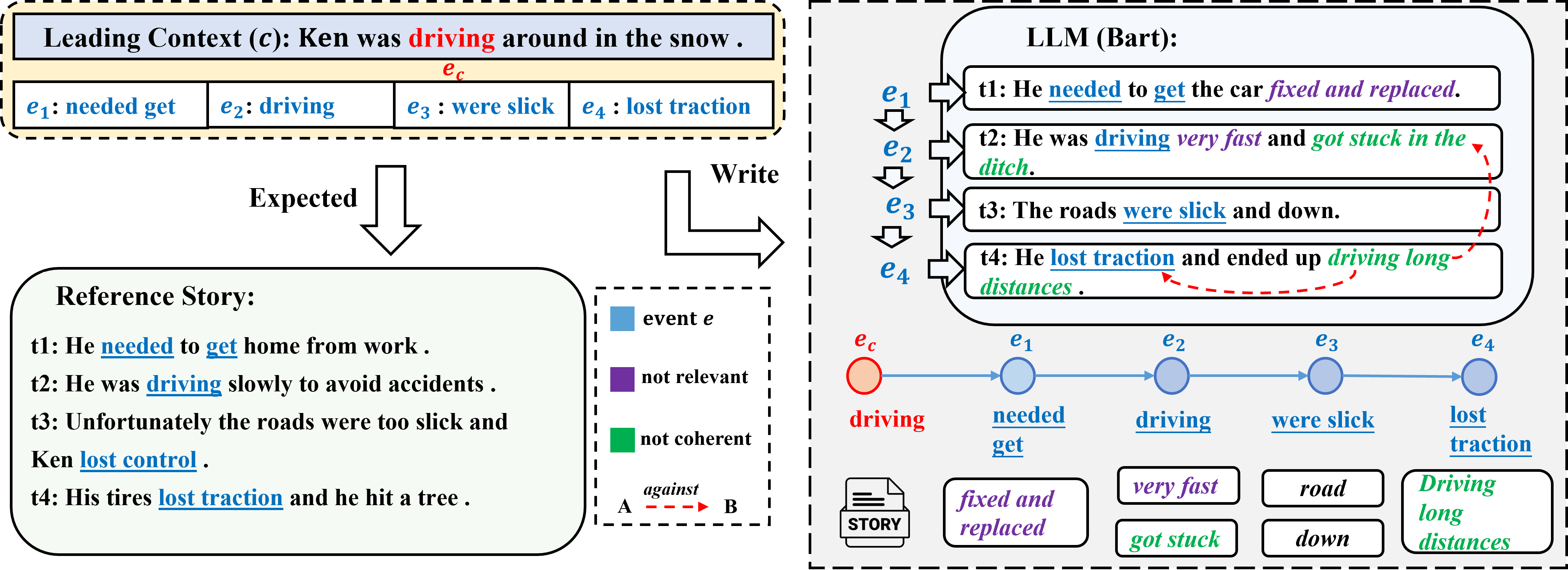}
\caption{Conditioned on leading context and reference events (extracted from reference stories), existing generation models still suffer from problems of relevance and coherence. For instance, we fine-tune BART \citep{lewis-etal-2020-bart} to generate stories. The leading context and reference text in this example are collected from ROC Stories \citep{mostafazadeh-etal-2016-corpus}. Some conflicts among them are observed and coloured. 
} 
\label{fig:examples}
\end{figure}

Despite notable advancements, existing methodologies still exhibit limitations in effectively leveraging planned events during story generation.
Conventionally, pre-trained language models (PLMs) such as BART~\citep{lewis-etal-2020-bart} are employed for generating narratives following event planning. However, as exemplified by the conflicts in Figure \ref{fig:examples}, while the individual sentences generated by BART may appear plausible, several issues arise when considering the coherence of the entire story. For instance, in a commonsense narrative, if a car is required to be \textit{``fixed and replaced''}, it is improbable for someone to then \textit{``drive around''}. Additionally, it is incongruous for Ken to drive the car very fast in the snow. Furthermore, if Ken\textit{``got stuck in the ditch''} or \textit{``lost traction''}, it is contradictory for him to then be \textit{"driving long distances"}. We postulate that these problems stem from the inadequacy of capturing contextual features while maintaining a coherent sequence of events. This is due to two primary reasons: (i) planned events often lack background information, such as the characteristics of Ken or the snowy setting, and (ii) training stories may contain identical events but differ in reference stories, leading to potential confusion during inference if the story-specific context is not considered.

Therefore, to address these challenges we propose \textbf{EtriCA} - a novel \textbf{E}vent-\textbf{Tri}ggered \textbf{C}ontext-\textbf{A}ware end-to-end framework for story generation. EtriCA enhances the generation process by effectively capturing contextual and event features from the input, surpassing the performance of state-of-the-art baseline models. Conventional generation models encounter difficulties in learning contextual representations while implicitly maintaining awareness of the event's state. This is primarily due to the feature disparities between events and contexts. Specifically, an event sequence (e.g. the verb), functioning as an abstract storyline, only contains schematic information pertaining to actions. On the other hand, the context encompasses story-specific details such as the scene and characters.

In order to comprehensively leverage the diverse range of features, our approach draws inspiration from previous research on information fusion \citep{abd2017analysing,chen-etal-2018-hybrid,xing-etal-2020-automatic,he-etal-2020-scene,you-etal-2020-hard, wang-etal-2021-fast,tang2022recent,zhanga2023cadge} to effectively encode heterogeneous features. To facilitate this encoding process, we employ a cross attention mechanism \citep{gheini2021cross,tang-etal-2022-EtriCA}. Our objective is to equip our model with a contextual understanding of the background as it unfolds during each event within a narrative structure. To accomplish this, we introduce a novel neural module designed to implicitly map contextual features to event features by employing information fusion techniques within their respective numeric vector spaces. This process, which we refer to as "contextualising events," allows our model to seamlessly integrate the relevant contextual information with the event representations. Leveraging the resulting contextualised event features, we employ an autoregressive decoder to dynamically generate stories, enabling the model to learn and unfold the contextualised events in a coherent manner. Furthermore, we introduce an auxiliary task of Sentence Similarity Prediction \citep{guan-etal-2021-long} to enhance the coherence between sentences driven by events. This additional task further strengthens the cohesion and logical flow within event-driven narratives, enhancing the overall quality and coherence of the generated stories.

Towards event-driven story generation, we introduce a novel task wherein stories are written based on a given leading context and event sequence. To enhance the event extraction framework proposed by \citet{chen-etal-2021-graphplan}, we leverage dependency parsing techniques to capture event-related roles from sentences, thus surpassing the limitations of heuristic rules. We also curate two datasets by pairing multi-sentence narratives from existing corpora with event sequences using our automated event extraction framework. Importantly, our task formulation can also benefit the study of controllable story generation, given the growing interest in neural generative frameworks centered around storylines \citep{xu-etal-2020-megatron, ghazarian-etal-2021-plot, chen-etal-2021-graphplan}. In order to evaluate the coherence and relevance automatically, we propose new automatic metrics based on semantic embeddings which can quantify the coherence and relevance of the generated stories. Through extensive experimentation, we demonstrate the superior performance of our proposed model, EtriCA, compared to baseline models in terms of fluency, coherence, and relevance metrics. 

Furthermore, to enhance the story writing capabilities of the proposed system, we employ a post-training framework that aims to improve the transferability and creative aptitude of EtriCA. In this paper, we will refer to this \textbf{K}nowledge \textbf{E}nhanced version as \textbf{KeEtriCA}. This framework enables EtriCA to adapt to a wider range of story genres, narrative structures, and plot dynamics. To achieve this, we utilise BookCorpus, a comprehensive story generation corpus, as the training dataset. BookCorpus consists of a vast collection of over 11,000 unique books encompassing diverse genres and authors, thereby encapsulating a wide spectrum of human knowledge, experiences, and narrative styles. The corpus covers a myriad of subjects, including both fiction and non-fiction genres such as mystery, romance, and science fiction, among others. The incorporation of such diverse content within BookCorpus facilitates the exploration and incorporation of various narrative techniques, fostering the development of innovative storytelling approaches. By leveraging the extensive textual data provided by BookCorpus, the language model of KeEtriCA is empowered to generate captivating and coherent stories across multiple genres, engaging readers and stimulating their imagination.

A range of experiments and in-depth analyses of our proposed framework are conducted by comparing it with the current state-of-the-art large-scale pre-trained models. The experimental results demonstrate that the stories produced by enhanced EtriCA exhibit superior performance in terms of relevance to the leading context and the given event sequences. This outcome highlights the advancement of our framework in achieving better controllability in story generation, surpassing single language models with increased hyper-parameters. As an extension of our work, we introduce \textbf{KeEtriCA}, which leverages a post-training framework to train models on the extensive BookCorpus dataset~\citep{zhu2015aligning} using our dependency-based event extraction method. This approach enhances the model's adaptability to a wider range of data samples. The contributions of our work can be summarised as follows: 

\begin{itemize}[noitemsep,nolistsep,leftmargin=*]
    \item We introduce a novel task in the domain of event-driven story generation. This task necessitates the generation model to compose narratives based on a specified initial context and a sequence of events. 
    \item We present an innovative method aimed at enhancing the existing event extraction framework. This enhancement is achieved by incorporating dependency parsing techniques. Furthermore, we provide annotated event sequences for two well-established datasets commonly used in our new task. 
    \item We propose a neural generation model \textbf{KeEtriCa}, which leverages the context and event sequence information with an enhanced cross-attention based feature capturing mechanism and sentence-level representation learning.
    \item We conduct a series of experiments and a comprehensive analysis to investigate the underlying characteristics contributing to writing a more fluent, relevant, and coherent story.
\end{itemize} 

\section{Related Work}

Prior to the ascent of deep learning techniques~\citep{tang2023terminology,huang-etal-2022-improving}, models for story generation predominantly relied on manual design principles, resulting in the production of rather simplistic sentences~\citep{mcintyre-lapata-2009-learning,woodsend-lapata-2010-automatic,mcintyre-lapata-2010-plot,huang-huang-2013-optimized,kybartas2016survey}. However, with the advent of neural story generation, end-to-end neural models, such as BART~\citep{lewis-etal-2020-bart} and GPT-2~\citep{radford2019language}, have gained widespread adoption as fundamental components for story composition~\citep{rashkin-etal-2020-plotmachines, guan-etal-2020-knowledge,goldfarb-tarrant-etal-2020-content,clark-smith-2021-choose}. Nonetheless, ensuring logical correctness becomes a challenging endeavor for straightforward Seq2Seq models as the generated text extends in length. This challenge has spurred recent research into the exploration of multi-step generations that seamlessly integrate neural models into conventional generative pipelines~\citep{guan-etal-2021-long}.

For instance, studies by \citet{yao2019plan,goldfarb-tarrant-etal-2020-content} and \citet{chen-etal-2021-graphplan} decompose the process of story generation into two distinct stages: planning (inputs-to-events) and writing (events-to-stories). They employ two separate neural generation models to facilitate learning at each stage.

In the planning stage, prior investigations~\citep{yao2019plan, rashkin-etal-2020-plotmachines, goldfarb-tarrant-etal-2020-content,jhamtani-berg-kirkpatrick-2020-narrative, ghazarian-etal-2021-plot} primarily concentrated on extracting event sequences from reference texts to serve as ground truths for plot planning. Neural models~\citep{radford2019language,lewis-etal-2020-bart} were then harnessed to predict events based on the initial context or titles. These events can be represented in various formats, including verbs or keywords. One straightforward approach, which aligns with our chosen method, involves the extraction of verbs to represent events~\citep{jhamtani-berg-kirkpatrick-2020-narrative, guan-etal-2020-knowledge, kong2021stylized}. However, the representation of verbs alone may fall short in preserving the integrity of information. For instance, the incorporation of semantic roles such as negation (e.g., "not") is pivotal for accurate comprehension. While heuristic rules have been employed by \citet{peng-roth-2016-two} and \citet{chen-etal-2021-graphplan} to include such semantic roles, it should be noted that these rules may not encompass all essential roles. Drawing inspiration from related work in open-domain event extraction~\citep{rusu-etal-2014-unsupervised,bjorne-salakoski-2018-biomedical,huang-etal-2018-zero,peng2021named}, we introduce an event extraction workflow based on dependency parsing. This approach allows us to capture the crucial components of verb phrases in sentences, which serve as events.

\section{Methodology} \label{sec:methodology}

\subsection{Task Formulation} \label{sec:task}
Within the domain of controllable story generation, we introduce a task that entails the creation of narratives through the effective fusion of a given initial context and a predetermined sequence of events. Our principal objective is to explore the harmonious integration of contextual information while upholding coherence with the provided event sequence, facilitated by neural generation models. To this end, we extend the context-aware story generation framework as originally proposed by \citep{guan-etal-2021-long}. We introduce an event sequence as a narrative guideline for each given initial context. Each input instance comprises a leading context, denoted as $C = {c_1, c_2, ..., c_n}$, which serves as the inaugural sentence of the narrative. Additionally, an event sequence is represented as $E = {e_1, e_2, ..., e_m}$ to delineate the storyline. Here, $c_i$ signifies the $i$-th token within the leading context, while $e_i$ corresponds to the $i$-th event, signifying the $i$-th sentence within the narrative. The ultimate output is a multi-sentence narrative denoted as $S = {s_1^1, s_2^1, ..., s_1^2, ..., s_n^m}$, with $s_j^i$ representing the $j$-th token within the $i$-th sentence of the narrative.

\subsection{Event Sequence Preparation} \label{event extraction}
\begin{figure*}[t]
\centering
\includegraphics[width=0.98\linewidth]{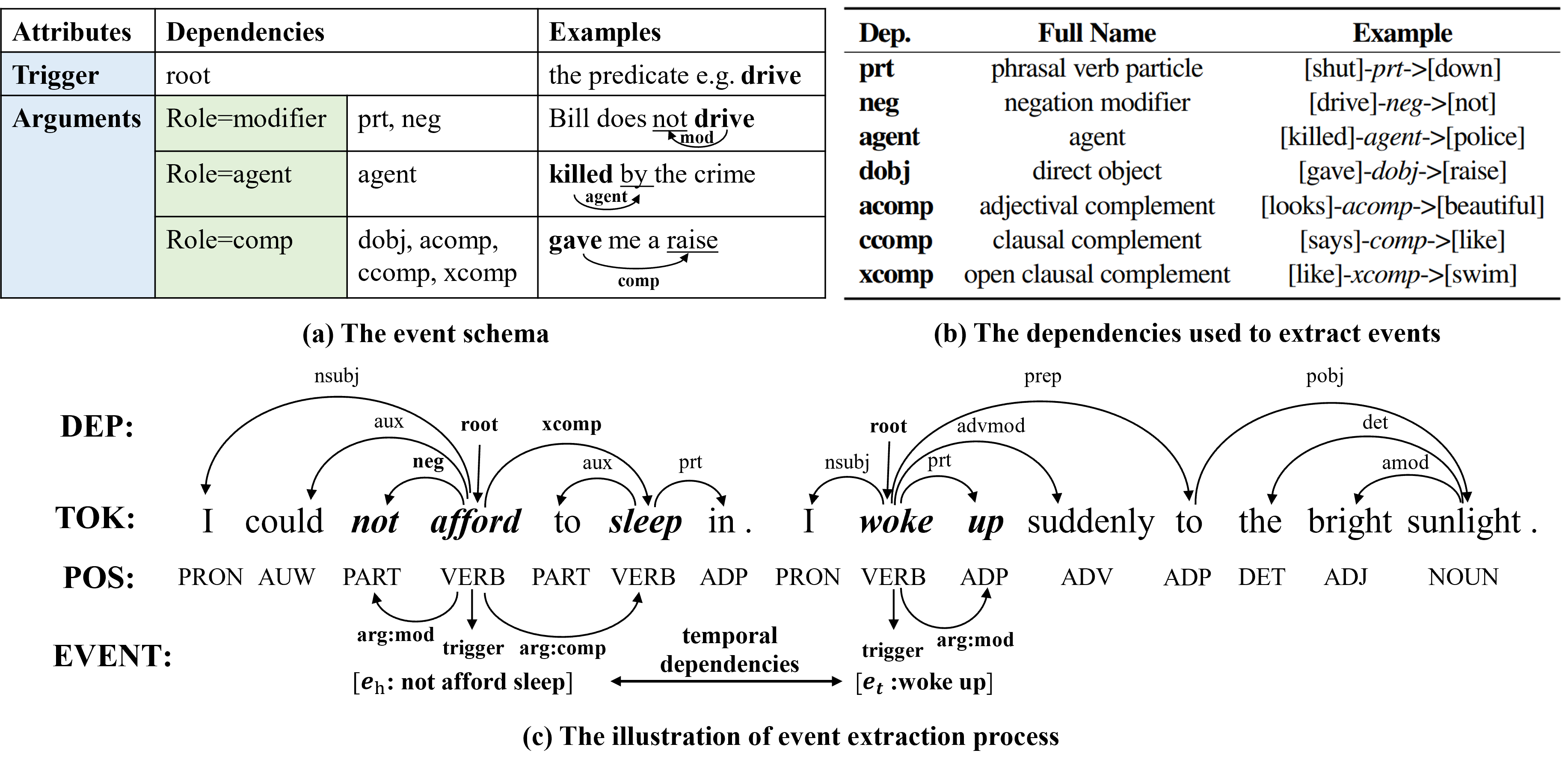}
\caption{The figure shows our event extraction process, where it includes three parts: (a) A table that delineates the schema of the extracted events. (b) An illustration of the relationship between sentence dependencies and the elements identified as our events. (c) An illustrative example that elucidates how events are extracted from input utterances. \textit{TOK} is the basic unit of a sentence. \textit{POS} is the part of speech, and \textit{DEP} stands for the dependencies between tokens. By parsing these dependencies, the event trigger assumes the responsibility of sieving out all significant roles necessary to represent a comprehensive action. Additionally, neighboring events that are extracted are considered to possess temporal relationships.}

\label{fig:event_process}
\end{figure*}

The details of our event extraction framework are presented in Figure \ref{fig:event_process}. We leverage the $spaCy$ library (\url{https://spacy.io/}) for sentence tokenisation and dependency parsing.

\paragraph{Event Schemas} Events serve as pivotal representations of significant changes or actions within a narrative. The function of an event schema is to encapsulate all pertinent roles associated with the action, distilling the core elements while filtering out superfluous details. Our methodology draws inspiration from the notable work of~\citet{rusu-etal-2014-unsupervised} and \citet{bjorne-salakoski-2018-biomedical}, who harnessed dependency parsing techniques to identify intricate word dependencies across diverse clauses. By leveraging the hierarchical structure of typed dependencies~\citep{de2008stanford}, we extract event mentions from sentences, resulting in substantially more informative and disambiguated events in contrast to the simplistic single-verb events employed in prior research~\citep{jhamtani-berg-kirkpatrick-2020-narrative, guan-etal-2020-knowledge, kong2021stylized}. An illustrative representation of the event schema is presented in \autoref{fig:event_process}.

Within \autoref{fig:event_process} (a), we illustrate the extraction of event arguments predicated on selected word dependencies. We also provide further elucidation on these dependencies and the specific linguistic roles they embody within a sentence. For a comprehensive understanding of these dependencies, we refer readers to the exhaustive study by \citet{de2008stanford}. In the construction of event schemas, a delicate equilibrium must be struck between generality and representational fidelity. The inclusion of additional dependencies has the potential to augment the informativeness of an event but may jeopardise its applicability across different contexts. For instance, the inclusion of the \textit{Subject} role (e.g., "I," "you," "Kent," etc.) can effectively characterise an event. However, given the varying characters featured in different stories, events extracted from one narrative may not seamlessly apply to another. For example, "Kent is driving" and "He is driving" convey an identical meaning, but if the subject "Kent" is extracted as an event role, it becomes intricate to predict the same event for a distinct narrative, thereby diminishing its generality. To mitigate this, we employ a set of stringent criteria to select key roles as event arguments, ensuring a harmonious balance between considerations of generality and representational fidelity.

\paragraph{Event Extraction} 
The process of event extraction entails the distillation of events from the text contained within the training dataset, encompassing both reference stories and leading contexts. Each event is represented as a set, comprehending the pivotal trigger and its associated arguments within a given sentence. Initially, we employ the $\mathit{spaCy}$ library for parsing word dependencies within sentences, subsequently annotating event triggers and their respective arguments based on these dependencies. Event $e$ embraces the attributes outlined in \autoref{fig:event_process} (b), with the event trigger, typically manifesting as the predicate, functioning as the root. Prior to integration into the encoders, the extracted events are serialised into text format to facilitate model ingestion.

As extant story datasets often lack reference storylines paired with reference stories, we have developed an event extractor capable of deriving event sequences from reference stories, which effectively serve as narrative foundations. In our methodology, events are embodied as verb phrases, with verbs serving as sentence anchors. Consequently, our core objective lies in the comprehensive extraction of all significant roles, herein referred to as "event arguments" that are intrinsically linked with the event trigger. The proximity of extracted events is indicative of temporal relationships.

To capture events bearing temporal relations from the training stories, we construct an event graph denoted as $G$. This graph stands as an isomorphic representation, featuring a singular event type and a singular relation type. We formally define $G$ as a data structure composed of triples in the format $e_h, r, e_t$. The workflow of the extraction process is elucidated as follows:

\begin{algorithm}[!ht]
\DontPrintSemicolon
  \KwInput{A story $ S $ comprising $ m $ sentences}
  \KwOutput{An Event Sequence $ E $ for $ S $ containing $ m $ event objects}
  Initialise $ E \leftarrow \varnothing $ and $ roles \leftarrow \{\mathit{trigger}, \mathit{mod}, \mathit{agent}, \mathit{comp}\} $
  \ForEach{$s^i$ in $S$}{
        Initialise $ e_i \leftarrow \varnothing $
        
        Normalise $ s^i $ and get dependencies $\mathit{dep}_i$ with $\mathit{spaCy}$
        
        Extract event trigger $t$ and position $p_t$ from $\mathit{dep}_i$
        
        $ e_i.\mathit{trigger} \leftarrow t$
        
          \ForEach{$\mathit{role}$ in $\mathit{role}$}{
              \If{$t \in \mathit{dep}_i.heads$ and $\mathit{role} \in \mathit{dep}_i.tails $}{
                  Extract $(\mathit{role}, p_{\mathit{r}})$ from $\mathit{dep}_i$
            
                  $ e_i.\mathit{role} \leftarrow (\mathit{role}, p_{\mathit{r}}) $
              }
      }
      $ e_i.\mathit{string} \leftarrow r \in \mathit{roles} $ aligned by $ p_r \uparrow$
      
      $E$ append $e_i$
  }
\caption{The algorithm of extracting Event Sequence $E$\label{alg:extract}}
\end{algorithm}

\begin{figure*}[ht]
\centering
\includegraphics[width=0.98\linewidth]{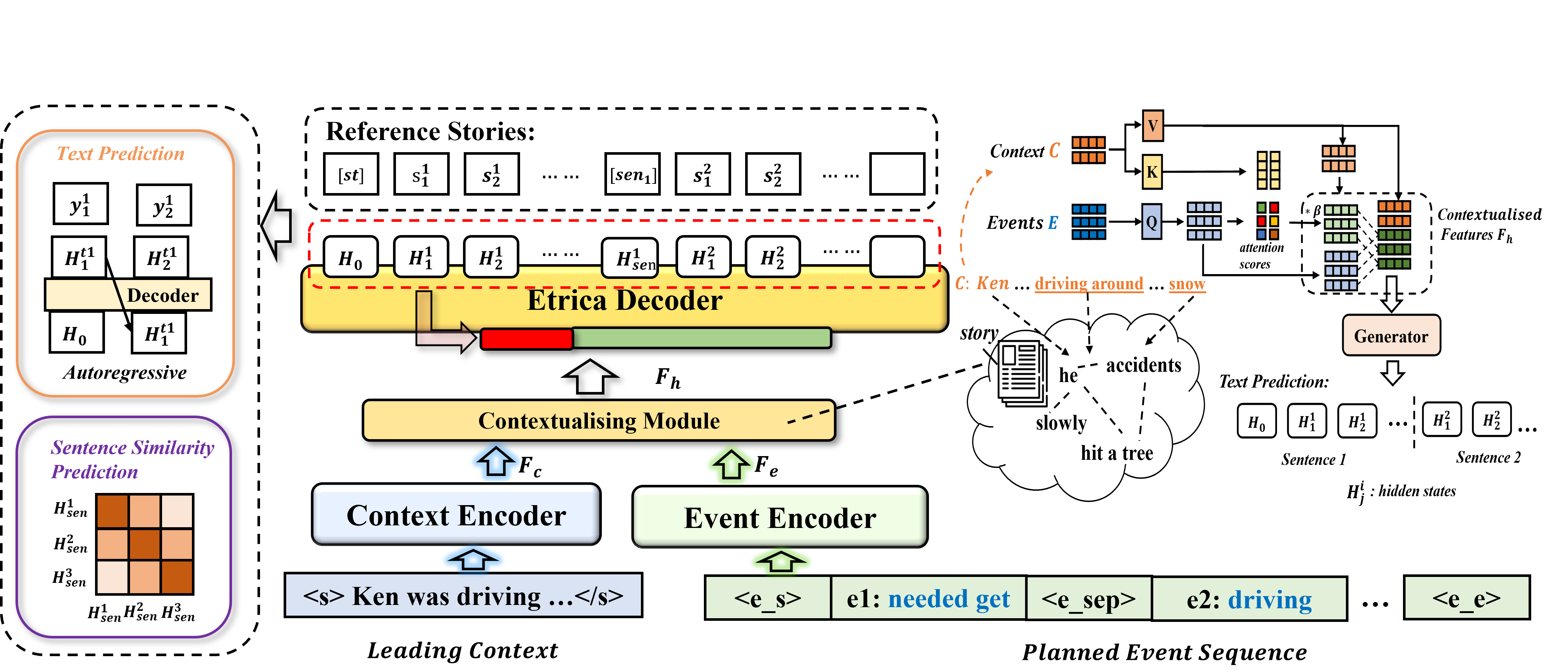}
\caption{The architecture of EtriCA. For a comprehensive understanding of the technical details, please refer to Section~\ref{neural-model}. During the training phase, in addition to the step-by-step prediction of text tokens ${y_1^1, \ldots, y_j^i}$, the decoder undergoes training to acquire sentence-level representations through the auxiliary task of similarity prediction, as depicted within the dotted box. This approach leverages representation learning, enabling neural models to acquire the proficiency required for generating stories akin to reference narratives while considering the provided leading context and planned event sequence.}
\label{fig:model}
\end{figure*}

\subsection{Neural Framework} \label{neural-model}

\subsubsection{The Main Architecture}\label{sec:main-arch}
Figure~\ref{fig:model} illustrates the main architectural components of our EtriCA framework for story generation.

\paragraph{Contextualised Features Representation}
Traditional language models commonly utilised in generation frameworks, such as transformers (Vaswani et al., 2017) or RNNs (Ghosh et al., 2017), are primarily designed for encoding natural language input text. As a result, the extracted events necessitate serialisation into plain text format. To accomplish this, we employ specialised tokens to demarcate the string format of events, as introduced in Section~\ref{event extraction}. For instance, we represent events as follows: ``<e\_s> needed get <e\_sep> ... <e\_e>'', where ``<e\_s>'',``<e\_sep>'', ``<e\_e>'' respectively signify the commencement, separation, and culmination of event planning.

During the encoding phase, the neural model receives inputs of $C$ and $E$, which exhibit distinct feature characteristics, as elaborated previously. Conventional end-to-end models often concatenate the embeddings of different inputs since neural encoders adeptly capture their features in a numeric vector space, frequently employing self-attention mechanisms. However, as the event sequence extends in length, the progressively growing concatenated embeddings may eclipse the influence of $C$. In response to this challenge, we employ two separate BART encoders~\citep{lewis-etal-2020-bart} to incorporate these features. Subsequently, we amalgamate these features using multi-head attention, calculated as follows:

\begin{align}
    & F_c = \text{Encoder}_c(C); F_e = \text{Encoder}_e(E) \\
    & Q_i = {W_i^Q} {F_e}, K_i = {W_i^K} {F_c}, V_i = {W_i^V} {F_c},\\
    & A_i = \text{softmax}(\frac{Q_iK_i^\mathrm{T}}{\sqrt{d_k}})V_i \\ 
    & F_{ca} = \text{Concat}(A_1, ..., A_m)W^M
\end{align}
In these equations, $Encoder_c$ and $Encoder_e$ inherit pre-trained parameters from BART but do not share trainable parameters during fine-tuning. $F_c$ and $F_e$ represent the features captured from $C$ and $E$, respectively. The subscript $i$ designates the $i$-th head of the attention scores, with a total of $m$ heads. $W_i^Q$, $W_i^K$, $W_i^V$, and $W^M$ are trainable parameters. The $i$-th head attention $A_i$ is computed as the attention-based weighted sum of the feature matrix. Ultimately, the obtained $F_{ca}$ represents the attention allocated to ongoing events, taking into account the contextual context.

To contextualise the input event features, we incrementally add $F_{ca}$ to the original event features $F_e$, forcing the neural models to learn the context gap between event sequences and stories. Mathematically, this is expressed as:
\begin{align}
    & F_{he} = F_e + \beta \odot F_{ca} \label{eq:beta} \\
    & F_h = \text{Concat}(F_c, F_{he})
\end{align}
where $\beta$ represents the scale factor applied to $F_{ca}$. $\beta \odot F_{ca}$ signifies the representation of the context gap, which is learned through residual mapping. The resulting feature vector $F_h$ is obtained by concatenating both the leading context features $F_c$ and the contextualised event features $F_{he}$. These combined features are then fed into a neural decoder to predict tokens and generate sentence representations.

\paragraph{Decoding and Sentence-level Fitting}\label{sec:sent}
In accordance with conventional generation systems, our approach leverages an auto-regressive decoder for the generation of story tokens $y_t$, following the equations below:
\begin{align}
    & H_t = \text{Decoder}(y_{<t}, F_h) \\
    & P(y_t|y_{<t}, X) = \text{softmax}(H_tW) \\
    & y_t \stackrel{sampling}{\longleftarrow} P(y_t|y_{<t}, F_h)
\end{align}
Here, $t$ signifies the time step, while $X$ denotes the input to the neural model. $H_t$ represents the hidden state at time $t$ within the decoder module. $H_t$ is computed by considering both the contextual and event-related information from $F_h$ as well as the previously predicted story tokens $y_{<t}$. $W$ is a trainable parameter, and $P(y_t|y_{<t}, F_h)$ is the probability distribution over the vocabulary, inclusive of special tokens. We employ a sampling strategy (e.g., $\mathit{argmax}$) to select the predicted token $y_t$.

In addition to token-level representations, we introduce an auxiliary task known as Sentence Similarity Prediction (Guan et al., 2021) to facilitate the acquisition of sentence-level representations and corresponding training methodologies. Given that an auto-regressive decoder predicts $y_t$ based on prior tokens $y_{<t}$, we enable the neural model to learn the generation of a specialized hidden state $H_{sep}^i$ corresponding to the position of a special token $[sep_i]$, where $i$ denotes the $i$-th sentence. We employ Sentence-BERT to obtain a numerical vector $F_i^{sent}$, encapsulating the features of individual sentences through representation learning. Subsequently, we enforce the similarity score $sim_{ij}^{s}$ between generated sentences to align with the similarities observed in reference stories, as computed in the equations below:
\begin{align}
    & F_i^{sent} = \textit{Sentence-Bert}(\{s_1^i, ..., s_n^i\})\label{eq:sent} \\
    & sim_{ij}^{s} = cosine(F_i^{sent}, F_j^{sent}) \\
    & u_{ij} = (H_{sep}^i)^\intercal W^{sep} H_{sep}^j \\
    & sim_{ij}^{y} = sigmoid(u_{ij} + u_{ji}) 
\end{align}
In these equations, $i$ and $j$ serve as indices for sentences, while $sim$ represents the similarity. $sim_{ij}^{s}$, the ground-truth similarity, is computed as the cosine similarity between the outputs of "Sentence-BERT." The variable $u_{ij}$ serves as an intermediate similarity measure derived from predicted sentence representations, and $W^{sep}$ represents a trainable parameter. To ensure that $sim_{ij}^{y}$ respects symmetry, considering both the $i$-to-$j$ and $j$-to-$i$ relationships, both $u_{ij}$ and $u_{ji}$ are included in the calculation.

\paragraph{Training and Inference}
In alignment with Figure~\ref{fig:model}, our neural model undergoes training to align with both token and sentence-level references, guided by the following objective functions:
\begin{align}
    & \mathcal{L}_{lm} = - \frac{1}{N}\sum_{t=1}^N logP(y_t|y_{<t}, X) \\
    & \mathcal{L}_{sent} = \frac{1}{m^2} \sum_{i=1}^m \sum_{j=1}^m (max|sim_{ij}^{s} - sim_{ij}^{y}|, \Delta) \label{eq:delta}\\
    & \mathcal{L}_{overall} = \mathcal{L}_{lm} + \lambda \mathcal{L}_{sent} \label{eq:lam}
\end{align}
Here, $\mathcal{L}{lm}$ characterises the cross-entropy loss of $P(y_t|y{<t}, F_h)$, encapsulating the token-level predictions. $\mathcal{L}{sent}$ encompasses the loss associated with predicted sentence similarities. Notably, $sim{ij}^{s}$ and $sim_{ij}^{y}$ denote the sentence similarities between the $i$-th and $j$-th sentences within a reference story and a generated story, respectively. The parameter $\lambda$ serves as an adjustable scale factor, with $\mathcal{L}{overall}$ representing the overarching loss function. Minimising $\mathcal{L}{overall}$ during training enables the neural model to generate stories that closely emulate human-like narratives. It is crucial to emphasise that the Sentence Similarity Prediction task is exclusively employed during the training phase. Consequently, during inference, the neural model produces stories without the presence of these special tokens.


\begin{figure*}[ht]
\centering
\includegraphics[width=0.98\linewidth]{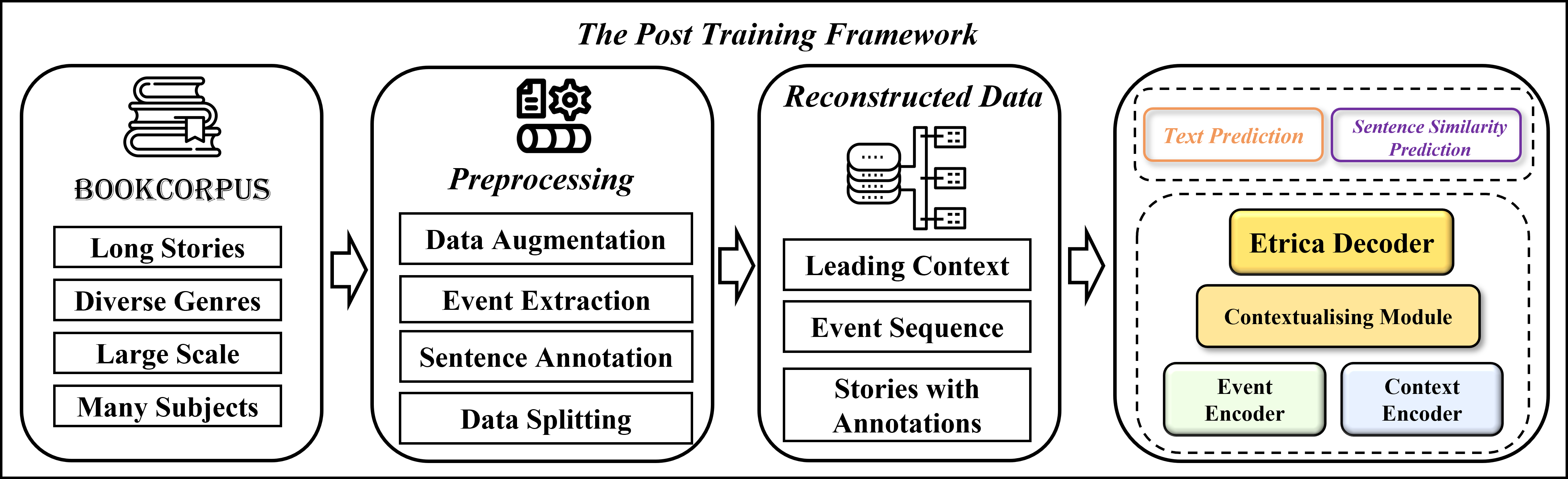}
\caption{An overview of the Post-Training Framework. This framework focuses on the preprocessing and reconstruction of the original corpus data to facilitate its utilisation by the EtriCA model.}
\label{fig:post}
\end{figure*}

\subsubsection{KeEtriCa: the Knowledge Enhanced Extension of EtriCA} \label{knowledge-enhancing} 

The aforementioned components constitute the core architecture referred as \textbf{EtriCA}.
EtriCA incorporates a contextualising module that takes into account both leading contexts and planned event sequences. However, due to the strict input requirements and limited size of the training datasets (ROC Stories and Writing Prompts), biases may be introduced during generation. These biases could hinder EtriCA's ability to comprehend given instructions and learn diverse generation paradigms. To overcome this limitation and enhance EtriCA's story writing proficiency across a broader range of genres, narrative structures, and story categories, we further improve its performance through post-training on a larger, more comprehensive, and diverse story corpus. The knowledge-enhanced EtriCA is then referred to as \textbf{KeEtriCa}.

To achieve this, we select BookCorpus~\citep{zhu2015aligning}, a substantial collection of freely available novels written by unpublished authors. This corpus consists of 11,038 books, comprising approximately 74 million sentences and 1 billion words, covering 16 distinct sub-genres. By leveraging this corpus, we expect EtriCA to perform better in few-shot or zero-shot settings, thus expanding its capabilities.

However, the raw corpus from BookCorpus cannot be directly employed for post-training in EtriCA due to its incorporation of heterogeneous features of leading contexts and events. Therefore, a preprocessing pipeline is employed to transform the raw stories. The pipeline consists of Data Augmentation, Event Extraction, Similarity Annotation, and Data Splitting stages. In the Data Augmentation stage, the raw stories are initially segmented with a maximum of 11 sentences. Each story is divided into a single input sentence and the subsequent ten sentences serving as the target output. Next, our proposed event extraction frameworks are applied to the target output, extracting the reference event sequence as the story plot. To represent the sentence-level features of the target story, we insert a special token ([sep\_i]) between neighboring sentences. Additionally, we employ Sentence-BERT~\citep{reimers-gurevych-2019-sentence} to collect representations of sentence embeddings, as described in Equation 1. These processed stories form the corpus for post-training and are subsequently split into training, evaluation, and test sets. The distribution of samples in these sets is as follows: 837,475 in the training set, 1,599 in the evaluation set, and 1,599 in the test set.

The reconstructed data is then fed into the main architecture, as described in \autoref{sec:main-arch}, for post-training. A total of 30 epochs of training are conducted using the same training settings outlined in \autoref{sec:implementation}. After the training process, the best checkpoint is selected, and the post-trained EtriCA is fine-tuned on each dataset of ROC Stories and Writing Prompts in the subsequent experiments.



\section{Experiment}

\subsection{Datasets}
We augment our dataset with additional event sequences, which are employed as benchmarks, through the annotation of two widely recognised datasets: ROC Stories (ROC) \citep{mostafazadeh-etal-2016-corpus} and Writing Prompts (WP) \citep{fan-etal-2018-hierarchical}. Our data preprocessing procedures align with those adopted in prior research \citep{xu-etal-2020-megatron,guan-etal-2021-long}. Specifically, we utilise the NLTK library~\citep{bird-loper-2004-nltk} to segment the narratives in both datasets into individual sentences. In the case of the ROC dataset, we perform delexicalisation by substituting all proper nouns with designated tokens, such as [MALE], [FEMALE], and [NEUTRAL]. In the case of the WP dataset, we curate data from the original development and test sets, retaining the initial eleven sentences from each narrative. This selection is necessitated by the extensive size and unbounded thematic diversity inherent to the original WP dataset. For both datasets, we designate the first sentence of each narrative as the leading context input, denoted as $C$, whereas the subsequent sentences constitute the reference story, denoted as $S$. This process enables us to compile an extended narrative dataset, WP (10 sentences), and a concise narrative dataset, ROC (4 sentences), for use in subsequent experiments. The event sequence, denoted as $E$, is derived from the reference story $S$, thereby representing the planned storyline that guides the story generation process. In terms of dataset statistics, the ROC dataset comprises Train/Dev/Test sets consisting of 88,344/4,908/4,909 narratives, respectively, while the WP dataset is partitioned into sets of 26,758/2,000/2,000 narratives.

\subsection{Baselines}

We conduct a comparative evaluation of our proposed model, EtriCA, against several state-of-the-art (SOTA) generation models as outlined below:
\begin{itemize}
    \item \textbf{P\&W} (Plan and Write) \citep{yao2019plan}: This model features a core architecture composed of Bidirectional Long Short-Term Memory (BiLSTM) with an attention mechanism~\citep{garg-etal-2019-jointly}. To ensure fair comparisons, we have improved upon the original code by substituting the static word embeddings with dynamic embeddings derived from the pre-trained BART model.

    \item  \textbf{GPT-2} \citep{radford2019language}: GPT-2 is a renowned auto-regressive generative model that has found extensive utility in prior research works \citep{rashkin-etal-2020-plotmachines, guan-etal-2020-knowledge,clark-smith-2021-choose}.
    
    \item \textbf{BART} \citep{lewis-etal-2020-bart}: BART is a composite model that amalgamates a BERT-like encoder \citep{devlin-etal-2019-bert} with a GPT-like decoder. It has demonstrated promising outcomes in a variety of natural language generation tasks~\citep{goldfarb-tarrant-etal-2020-content,clark-smith-2021-choose}.
    
    \item  \textbf{HINT} \citep{guan-etal-2021-long}: HINT presently represents the state-of-the-art framework for context-aware story generation. It elevates coherence and relevance through supplementary sentence-level and discourse-level training.
\end{itemize}
These models collectively serve as robust baselines for the assessment of EtriCA's performance in the domain of context-aware story generation.

\subsection{Implementation Details} \label{sec:implementation}
The primary contribution of our generation model lies in its contextualising module, which can be seamlessly integrated into various encoder-decoder frameworks. To harness the notable performance achieved in prior research~\citep{goldfarb-tarrant-etal-2020-content,guan-etal-2021-long}, we have adopted the encoders and decoders from the BART framework~\citep{lewis-etal-2020-bart} as the foundation for our neural generation model. For the fine-tuning of our generation model, we have employed a publicly available BART checkpoint\footnote{\url{https://huggingface.co/facebook/bart-base}} from the Huggingface model hub. To ensure reproducibility, we have maintained the fixed random seed of $42$ throughout our experiments. All code implementations have been carried out using the PyTorch library and trained using the PyTorch Lightning framework. The hyper-parameters are set as follows: the residual scale factor, denoted as $\beta$ in Equation \ref{eq:beta}, is set to $0.1$, the margin, represented as $\Delta$ in Equation \ref{eq:delta}, is set to $0.1$, and the scale factor, indicated as $\lambda$ in Equation \ref{eq:lam}, is set to $0.1$. Furthermore, certain parameters are learned during training on our datasets. In our framework, both the encoders and the decoder are structured with six hidden layers and implement a $12$-head attention mechanism. The shared embedding layer comprises a vocabulary of up to 50,625 tokens, encompassing Byte-Pair Encoding~\citep{radford2019language}  and additional special tokens mentioned in \S \ref{sec:methodology}.

Our experiments have been conducted using multiple GPUs, specifically RTX A4000s, on a cloud platform. To ensure reproducibility, we have maintained a fixed random seed of $42$. The training process has been implemented within the PyTorch Lightning framework, offering various APIs that simplify the engineering process. The specific training parameters include a batch size of $64$, a learning rate of $8e-5$, a maximum source length of $1024$, and an Adam optimiser~\citep{kingma2014adam} with an epsilon value of $1e-8$. The training process consists of five epochs, with the best-performing checkpoint determined based on the loss metric, aiming for the lowest loss value. It is essential to note that EtriCA necessitates two separate encoders for encoding context (natural language) and events (concatenated serialised events) individually. However, the encoder from the public BART checkpoint has been pre-trained solely on natural language text. To enhance the learning of event features within the event encoder, we have initially trained a BART model on stories that incorporate both the context and planned events. Subsequently, we have transferred the pre-trained encoder parameters to the event encoder of EtriCA. During the inference phase for evaluation and testing, we have employed the nucleus sampling strategy~\citep{holtzman2019curious} for text generation. Additionally, we have adjusted the batch size to $15$ during inference, as nucleus sampling requires a larger memory footprint.

\subsection{Evaluation Metrics}
\textbf{Perplexity (PPL)} is a metric that quantifies the uncertainty of predicted tokens generated by neural models. 
To assess the quality of generated stories, we employ several reference metrics, including \textbf{ROUGE-n (R-n)} \citep{lin-2004-rouge} and \textbf{BLEU-n (B-n)} \citep{papineni-etal-2002-bleu}.  ROUGE-n measures the coverage rate between the generated stories and the referenced stories, where $n$ denotes the n-gram order. Similarly, BLEU-n computes the n-gram overlaps between the generated stories and the references. In addition to these reference metrics, we utilise several unreferenced metrics to evaluate the quality of generated stories. \textbf{Lexical Repetition-n (LR-n)}~\citep{shao-etal-2019-long} is a metric that quantifies the percentage of generated stories containing a 4-gram that is repeated at least $n$ times. \textbf{Distinction-n (D-n)}~\citep{li-etal-2016-diversity}  is another unreferenced metric that measures the distinction of stories by calculating the ratio of distinct n-grams to all generated n-grams.

\textbf{Intra-story Repetition} (Yao et al., 2019) quantifies sentence repetition within a story by measuring trigram overlaps. \textbf{Intra-story Coherence and Relevance} (Xu et al., 2018), originally developed for dialogue evaluation, calculates sentence-level coherence and relevance based on cosine similarity between semantic embeddings.\footnote{We employ GloVe Vectors for this purpose from the Stanford NLP group: \url{https://nlp.stanford.edu/projects/glove/}} In our study, we adapt this approach to assess the relatedness between consecutive generated sentences as intra-story coherence and the relatedness between the leading context and the story sentence as intra-story relevance.\footnote{Although originally an unsupervised metric designed for conversations, we repurpose it to evaluate our generated stories by applying it to the sentences within a story. The code used is sourced from the repository available at \url{https://github.com/tonywenuon/dialog-coherence-metric}.} \textbf{Intra-story Aggregate Metrics} encompass repetition, coherence, and relevance, which are obtained by calculating the mean of the corresponding sentence-level metrics.

\subsection{Automatic Evaluation for EtriCA}

\begin{figure}[tb]
\centering
\includegraphics[width=0.45\linewidth]{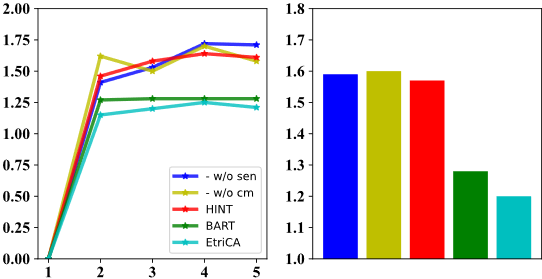}
\caption{The results of Intra-Story Repetitions and Aggregate Scores in ROC Dataset Narratives
The curves graphically represent the extent of intra-story repetition for each sentence within a narrative, with the leading context serving as the initial sentence. Meanwhile, the histograms portray the cumulative scores for intra-story repetitions across the sentences in the narrative.}
\label{fig:intra-inter-rept}
\end{figure}

The automatic evaluation results on the short story dataset ROC and the long story dataset WP are presented in Table~\ref{tab:referenced}. EtriCA surpasses all baselines across all the reference metrics for both datasets. Notably, compared to the strongest baselines, BART and HINT, our model achieves a perplexity reduction of $15\%$ on ROC and $25\%$ on WP. EtriCA also outperforms other baselines in terms of BLEU and ROUGE metrics, indicating that it generates stories that closely resemble human-written reference stories. We also examine the repetition and diversity of the generated stories. EtriCA demonstrates strong performance in terms of both lexical repetition (LR-2) and diversity (D-4), either achieving the best performance or being on par with the best-performing baseline models.
To gain further insights into how our model performs in writing along with the planned events, we adopt the approach of \citet{yao2019plan} to examine the intra-story repetitions for each generated sentence, as depicted in ~\autoref{fig:intra-inter-rept}. The results consistently show that EtriCA outperforms the baselines in both sentence-level and story-level (i.e., aggregated) repetition scores, indicating its superior performance in event-triggered story writing.

In addition,  the ablation study demonstrates the importance of both context and event features in enhancing the generation process. The performance of the \textit{- w/o leading} and \textit{- w/o events} variants indicate that the features present in these two types of inputs are complementary to each other and both are crucial for generating high-quality stories. Therefore, effectively incorporating both features becomes essential for improving story writing ability. When EtriCA does not implement our contextualising module (abbr. cm), all metrics significantly decrease, and some metrics even fall below those of BART${l+e}$ and HINT${l+e}$. This observation suggests that our contextualising module can more effectively fuse heterogeneous features and generate a richer semantic representation for subsequent story writing. Similarly, sentence-level representations also lead to improvements in most metrics, although not to the same extent as the contextualising module. We hypothesise that the contextualising module significantly reduces the gap between event sequences and stories (each event is paired with each sentence), making the improvement from sentence-level representations less pronounced. This hypothesis is further supported by subsequent experiments.

\begin{table*}[ht]
\centering \small
\resizebox{1.0\linewidth}{!}{
\begin{tabular}{l|cccccccc|cccccccc}
\toprule[2pt]
\multirow{2}{*}{\textbf{Models}} & \multicolumn{8}{c}{\textbf{ROC Stories}} & \multicolumn{8}{c}{\textbf{Writing Prompts}} \\
  & \textbf{PPL}$ \downarrow $ & \textbf{R-1}$ \uparrow $ & \textbf{R-2}$ \uparrow $ & \textbf{R-L}$ \uparrow $ & \textbf{B-1}$ \uparrow $ & \textbf{B-2}$ \uparrow $  &  \textbf{LR-2}$ \downarrow $ & \textbf{D-4}$ \uparrow $  
  & \textbf{PPL}$ \downarrow $ & \textbf{R-1}$ \uparrow $ & \textbf{R-2}$ \uparrow $ & \textbf{R-L}$ \uparrow $ & \textbf{B-1}$ \uparrow $ & \textbf{B-2}$ \uparrow $  &  \textbf{LR-2}$ \downarrow $ & \textbf{D-4}$ \uparrow $  \\
\midrule

\textbf{P\&W$_{l+e}$} & 6.22 & 22.82 & 2.65 & 15.90 & 0.297 & 0.150 & 0.297 & 0.773        & 16.47 & 23.49 & 1.74 & 12.17 & 0.259 & 0.086 & 0.443 & 0.834 \\
\textbf{GPT-2$_{l+e}$} & 10.02 & 29.85 & 6.45 & 20.58 & 0.347 & 0.201 & 0.528 & 0.675     & 49.48 & 18.59 & 2.18 & 10.38 & 0.130 & 0.051 & 0.760 & 0.684 \\
\textbf{BART$_{l+e}$}  & 3.39 & 48.74 & 21.95 & 40.69 & 0.505 & 0.351 & 0.245 & \textbf{0.804}       & 10.84 & 37.19 & 8.14 & 22.73 & 0.351 & 0.174 & 0.378 & 0.894 \\

\textbf{HINT$_{l+e}$}  & 3.97 & 46.71 & 20.81 & 37.21 & 0.488 & 0.337 & 0.264 & 0.734      & 14.45 & 38.86 & 8.98 & 23.06 & 0.373 & 0.190 & \textbf{0.338} & 0.855 \\
\midrule[1pt]
\textbf{EtriCA (ours)} & \textbf{2.88} & \textbf{49.29} & \textbf{22.59} & \textbf{41.43} & 0.506 & 0.354 & \textbf{0.244} & 0.799       & \textbf{8.11} & \textbf{39.90} & \textbf{9.65} & \textbf{25.21} & \textbf{0.387} & \textbf{0.202} & 0.359 & 0.889 \\ 
\textbf{- w/o sen} & 3.33 & 49.18 & 22.39 & 41.09 & \textbf{0.512} & \textbf{0.359} & 0.286 & 0.794       & 9.88 & 39.88 & 9.37 & 24.86 & 0.385 & 0.199 & 0.343 & \textbf{0.900} \\
\textbf{- w/o cm} & 2.97 & 48.53 & 21.55 & 40.34 & 0.499 & 0.345 & 0.245 & 0.800       & 9.15 & 36.08 & 7.55 & 21.01 & 0.356 & 0.175 & 0.514 & 0.827 \\ 
\textbf{- w/o leading} & 3.24 & 42.55 & 17.21 & 35.90 & 0.450 & 0.287 & 0.260 & 0.795        & 9.37 & 35.46 & 7.22 & 20.69 & 0.357 & 0.172 & 0.517 & 0.892 \\ 
\textbf{- w/o events} & 4.50 & 24.51 & 2.70 & 16.86 & 0.311 & 0.156 & 0.245 & 0.792       & 12.77 & 23.77 & 1.89 & 12.26 & 0.263 & 0.089 & 0.412 & 0.850 \\
\midrule[1pt]
\textbf{Golden} - & - & - & - & - & - & - & 0.048 & 0.906   & - & - & - & - & - & - & 0.286 & 0.950 \\
\bottomrule[2pt]
\end{tabular}
}
\caption{Automatic evaluation on ROC and WP datasets. 
The optimal performance in each category is indicated in bold font. The symbols $\uparrow$ and $\downarrow$ signify that higher or lower scores are desirable, respectively. The subscript $_{l+e}$ designates that the model input is a concatenation of the leading context and event sequence. The notations \textbf{w/o sen}, \textbf{w/o cm}, \textbf{w/o leading}, and \textbf{w/o events} signify the removal of the auxiliary task of sentence similarity prediction (the contextualizing module), the exclusion of the leading features, the omission of the leading context, and the absence of event features, respectively. The term \textbf{Golden} denotes the reference stories present in the datasets.}
\label{tab:referenced} \label{tab:unreferenced}
\end{table*}

\begin{figure*}[tb]
\centering
\includegraphics[width=0.7\columnwidth]{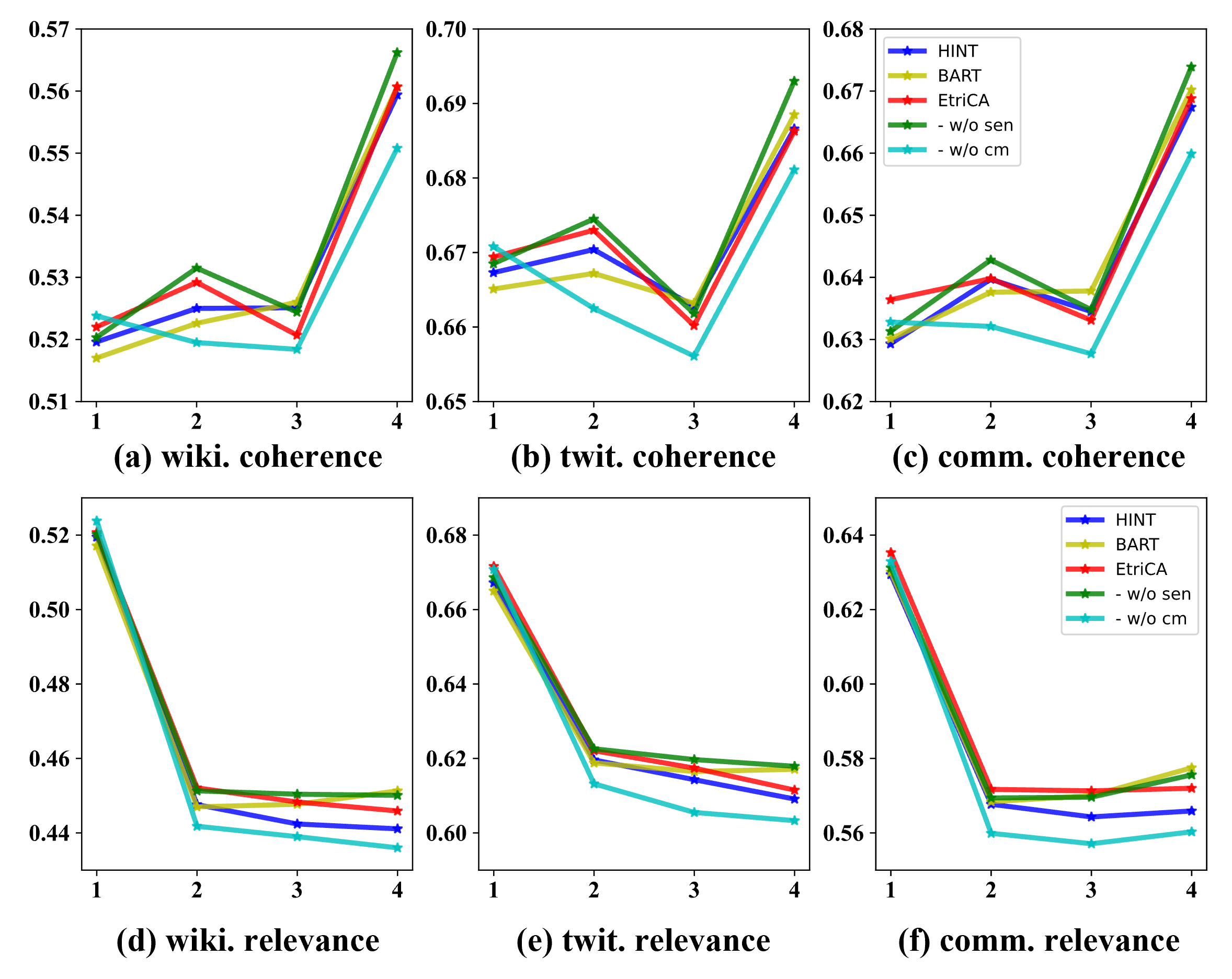}
\caption{The results of intra-story coherence and relevance on the ROC dataset.}
\label{fig:sent-coh-rel}
\end{figure*}

Furthermore, in order to gain further insights into the coherence and relevance of our generated stories, we present additional experimental results for an in-depth analysis. It is important to note that due to the nature of the WP dataset, which contains a substantial number of short and conversational sentences that lack meaningful analysis, we were unable to conduct intra-story analysis (repetition, coherence, relevance) on this dataset. As a result, we focused our intra-story experiments solely on the ROC dataset. To ensure a fair comparison, we selected the two strongest baselines based on previous experimental results. As demonstrated in \autoref{tab:coherence_and_relevance_roc}, our approach consistently outperforms the baselines in terms of intra-story coherence and relevance. These results highlight the effectiveness of our contextualising module in capturing relevant features from both the context and events, thereby enhancing the logical connectedness between story sentences and the overall coherence between the story and its input. Additionally, the ablation results reveal that the performance of EtriCA and the model without sentence-level representations (denoted as ``- w/o sen'') are very similar. This finding further supports our hypothesis that the feature capturing mechanism of the contextualising module partially replaces the role of sentence-level representations. To visually illustrate the superiority of our model, Figure~\ref{fig:sent-coh-rel} displays the performance comparison for coherence and relevance between our model and the baselines. It is evident that our model consistently outperforms the baselines, reaffirming its ability to capture relevant features and generate stories that are more closely related to the provided events and context.

\begin{table}[tb]
\centering
\resizebox{0.6\linewidth}{!}{
\begin{tabular}{l|ccc|ccc}
\toprule[1pt]
\multirow{2}{*}{\textbf{Models}} & \multicolumn{3}{c}{\textbf{Coherence}} & \multicolumn{3}{c}{\textbf{Relevance}} \\
  & \textbf{wiki.} & \textbf{twit.}& \textbf{comm.}  & \textbf{wiki.} & \textbf{twit.}& \textbf{comm.}  \\
\midrule
\textbf{BART$_{l+e}$}  & 0.4658 & 0.6293 & 0.5865      & 0.5316 & 0.6710 & 0.6439    \\
\textbf{HINT$_{l+e}$}  & 0.4627 & 0.6276 & 0.5818      & 0.5323 & 0.6718 & 0.6427  \\
\midrule
\textbf{EtriCA} & 0.4667 & 0.6306 & \textbf{0.5876}     & 0.5332 & 0.6722 & 0.6445  \\ 
\textbf{- w/o sen} & \textbf{0.4680} & \textbf{0.6322} & 0.5864     & \textbf{0.5356} & \textbf{0.6745} & \textbf{0.6457}    \\
\textbf{- w/o cm} & 0.4602 & 0.6232 & 0.5775       & 0.5281 & 0.6676 & 0.6381   \\ 
\midrule
\textbf{Golden} & 0.6631 & 0.7996 & 0.8298        & 0.6610 & 0.7997 & 0.8265    \\ 
\bottomrule[1pt]
\end{tabular}
}
\caption{The results of aggregate scores of intra-story coherence and relevance for the ROC dataset, which are calculated based on semantic embeddings. \textbf{wiki.}, \textbf{twit.}, \textbf{comm.} denotes the GloVe embeddings of ``Wikipedia 2014 + Gigaword 5 (6B tokens)'', ``Twitter (2B tweets, 27B tokens)'', and ``Common Crawl (42B tokens)'', respectively.}
\label{tab:coherence_and_relevance_roc}
\end{table}

\subsection{Human Evaluation For EtriCA} 
\label{human-evaluation}

\begin{table}[ht]
\centering
\resizebox{0.8\linewidth}{!}{
\begin{tabular}{r|lc|c|lc|c|lc|c}
\toprule[1pt]
\multirow{2}{*}{\textbf{Choices(\%)}} & \multicolumn{3}{c}{\textbf{Etri.} vs. \textbf{\textit{w/o cm}}}  & \multicolumn{3}{c}{\textbf{Etri.} vs. \textbf{BART$_{l+e}$}} & \multicolumn{3}{c}{\textbf{Etri.} vs. \textbf{HINT$_{l+e}$}}  \\
\cline{2-10} 
& \textbf{Etri.} & \textbf{\textit{w/o cm}} & \textbf{Kappa} & \textbf{Etri.} & \textbf{BART$_{l+e}$} & \textbf{Kappa} & \textbf{Etri.} & \textbf{HINT$_{l+e}$} & \textbf{Kappa}   \\
\midrule
\textbf{Fluency}  & \textbf{36.1}$^{**}$ & 18.0 & 55.3 & \textbf{33.6}$^{**}$ & 16.4 & 56.2  & \textbf{32.3}$^{**}$ & 17.3 & 55.4  \\
\textbf{Coherence}  & \textbf{40.2}$^{**}$ & 22.7 & 48.9 & \textbf{32.8}$^{*}$ & 19.1 & 48.1 & \textbf{35.3}$^{*}$ & 21.8 & 58.3 \\
\textbf{Relevance}  & \textbf{23.2} & 21.7 & 48.5 & \textbf{16.8} & 9.8 & 45.1 & \textbf{14.9} & 8.5 & 50.9  \\

\bottomrule[1pt]
\end{tabular}
}
\caption{Human Assessment on the ROC Dataset. The provided scores signify the percentage of instances in which our model was preferred over another model in pair-wise comparisons. To gauge inter-annotator agreement, the Fleiss' Kappa coefficient \citep{fleiss1971measuring} was employed. Notably, all of our findings have attained a level of moderate agreement. The symbols $*$ and $**$ denote significance at the p<0.05 and p<0.01 levels, respectively, as determined by a sign test.}
\label{tab:human_eval}
\end{table}

We conducted human evaluation using pairwise comparisons involving two competitive baselines and an ablated model lacking our proposed contextualising module. A total of 150 stories were randomly sampled from the ROC Stories test dataset for this evaluation. Three evaluators were invited to assess which generated story was superior (Win/Lose/Tie) based on three aspects: (\romannumeral1) \textbf{Fluency}, which considers the quality of each sentence, in isolation, from a linguistic perspective, including grammatical correctness and accurate representations of semantic meaning; (\romannumeral2) \textbf{Coherence}, which measures the logical connectedness between story sentences; and (\romannumeral3) \textbf{Relevance}, which evaluates the contextual relevance between the generated stories and the leading contexts. The final results were determined through majority voting based on the human assessments.

Table~\ref{tab:human_eval} presents the results of the human evaluation, where EtriCA outperforms the state-of-the-art (SOTA) baselines in terms of fluency, coherence, and relevance. Since all generation models exhibit relatively minimal deviations from the given input, they perform well in terms of relevance, resulting in fewer differences in this aspect. However, the improvements in fluency and coherence are highly significant, highlighting the advantages of our contextualising module in capturing high-level features from both the context and event sequences.

\subsection{Case Study For EtriCA} \label{sec:case-study}
\begin{table*}[ht]
\centering
\resizebox{0.95\linewidth}{!}{
\begin{tabular}{l|ll}
\toprule[2pt]
 \multirow{2}{*}{\textbf{Input}} & \textbf{Leading Context} & [MALE] had lost his dog over a month ago .\\
& \textbf{Event Sequence} & missed dog $\rightarrow$ notices something $\rightarrow$ sees dog $\rightarrow$ turns out be \\
\midrule[1pt]
\textbf{P\&W$_{l+e}$} & \multicolumn{2}{p{0.9\textwidth}}{He wished he could live with his friend. He 'd run in them all the time. But one day, he woke up exhausted. He went to the doctor with his best friend.} \\
\midrule
\textbf{GPT-2$_{l+e}$} & \multicolumn{2}{p{0.9\textwidth}}{[MALE] was only a parent at the time. [MALE] \textbf{notices} the dog and he lets it go. He notices the dog has been moved and so he notices what happened. [MALE] then realises that it is a bad dog and there is something wrong with his life.} \\
\midrule
\textbf{BART$_{l+e}$}  & \multicolumn{2}{p{0.9\textwidth}}{He \textbf{missed} his \textbf{dog} for a whole month. One day he \textbf{notices something} moving and is startled. He \textbf{sees} the \textbf{dog} on the floor. It \textbf{turns out} to \textbf{be} a squirrel.} \\
\midrule
\textbf{HINT$_{l+e}$}  & \multicolumn{2}{p{0.9\textwidth}}{One day [MALE] \textbf{missed} his \textbf{dog}. He \textbf{notices something} about her name on the dog's tag. [MALE] \textbf{sees} the \textbf{dog} in the tags. It \textbf{turns out} it could \textbf{be} a dog from the police department.}  \\
\midrule[2pt]
\textbf{KeEtriCA} & \multicolumn{2}{p{0.9\textwidth}}{He \textbf{missed} his \textbf{dog} badly. He \textbf{notices something} strange on the curb. He \textbf{sees} the \textbf{dog} outside. It \textbf{turns out} to \textbf{be} a stray dog.}  \\
\midrule
\textbf{- w/o sen} & \multicolumn{2}{p{0.9\textwidth}}{He had \textbf{missed} his \textbf{dog} so much that he had to search for him. As he was searching, he \textbf{notices something} about a dog. He \textbf{sees} the \textbf{dog} with a bag. It \textbf{turns out} to \textbf{be} a stray, a wad of dog spray.}  \\
\midrule
\textbf{- w/o cm} & \multicolumn{2}{p{0.9\textwidth}}{[MALE] \textbf{missed} his \textbf{dog} this summer. He \textbf{notices something} on his neighbor's wall about a house. [MALE] notices the dog was very sad. it \textbf{turns out} that there must \textbf{be} a really sad day next time.} \\ 
\midrule
\textbf{- w/o leading} & \multicolumn{2}{p{0.9\textwidth}}{He \textbf{missed} his \textbf{dog}. [MALE] \textbf{notices something} in the area. he \textbf{sees} a \textbf{dog}. it \textbf{turns out} to \textbf{be} a black dog.} \\ 
\midrule
\textbf{- w/o events} & \multicolumn{2}{p{0.9\textwidth}}{He was devastated by the loss. He decided to pull a long string of nail polish. He found a couple of old nail polish cans that were very old. His dog enjoyed his touches.}\\
\midrule
\textbf{Golden} & \multicolumn{2}{p{0.9\textwidth}}{He \textbf{missed} his \textbf{dog} very much. One evening while mowing the lawn he \textbf{notices something}. He \textbf{sees} a \textbf{dog} in the street that looked like his lost dog. It \textbf{turns out} to \textbf{be} his lost dog who had returned home.} \\
\bottomrule[2pt]
\end{tabular}
}
\caption{A case study of generated stories conditioned with a leading context and an event sequence collected from \textbf{ROC Stories}. \textit{[MALE]}, \textit{[FEMALE]}, and \textit{[NEUTRAL]} are the specital tokens to replace names in story. The highlighted bold words denote the events corresponding to the given event sequence.}
\label{tab:case_study}
\end{table*}

As depicted in Table~\ref{tab:case_study}, EtriCA outperforms the baseline models in generating stories that exhibit better contextual relevance and overall quality. The strong baseline models, namely \textbf{BART${l+e}$} and \textbf{HINT${l+e}$}, demonstrate proficiency in adhering to the planned event sequences and maintaining a satisfactory level of fluency. However, they fall short in crafting coherent stories that establish logical connections with the ongoing circumstances. For example, the sentence "It turns out to be a squirrel." may utilise the event phrase "turns out to be," but it lacks any relevance to the main topic of the missing dog and lacks coherence with the preceding sentences.

Regarding the results obtained from the ablation study, we observe the significance of different components within the overall generation model. When the planned event sequence is omitted (\textbf{- w/o events}), it becomes exceedingly challenging for a neural model to generate a coherent story, as supported by previous research \citep{yao2019plan}. Similarly, when the leading context is absent (\textbf{- w/o leading}), neural models encounter difficulties in unfolding the planned events due to the absence of a conceptual understanding of the "topic" of the story, leading to potential confusion. Without the inclusion of the contextualising module (\textbf{- w/o cm}), the neural model struggles to effectively process the diverse features derived from both the context and events, impeding the overall performance.


\begin{table*}[ht]
\centering \small
\resizebox{1.0\linewidth}{!}{
\begin{tabular}{l|cccccccc|cccccccc}
\toprule[2pt]
\multirow{2}{*}{\textbf{Models}} & \multicolumn{8}{c}{\textbf{ROC Stories}} & \multicolumn{8}{c}{\textbf{Writing Prompts}} \\
  & \textbf{PPL}$ \downarrow $ & \textbf{R-1}$ \uparrow $ & \textbf{R-2}$ \uparrow $ & \textbf{R-L}$ \uparrow $ & \textbf{B-1}$ \uparrow $ & \textbf{B-2}$ \uparrow $  &  \textbf{LR-2}$ \downarrow $ & \textbf{D-4}$ \uparrow $  
  & \textbf{PPL}$ \downarrow $ & \textbf{R-1}$ \uparrow $ & \textbf{R-2}$ \uparrow $ & \textbf{R-L}$ \uparrow $ & \textbf{B-1}$ \uparrow $ & \textbf{B-2}$ \uparrow $  &  \textbf{LR-2}$ \downarrow $ & \textbf{D-4}$ \uparrow $  \\
\midrule

\textbf{ChatGLM} & - & 40.77 & 13.15 & 28.69 & 0.301 & 0.180 & \textbf{0.120} & 0.791     & - & 30.24 & 4.89 & 16.12 & 0.196 & 0.078 & 0.615 & 0.727  \\ 

\textbf{ChatGPT} & - & 43.22 & 15.11 & 31.44 & 0.321 & 0.203 & 0.147 & \textbf{0.899}     & - & 35.67 &  6.79 & 19.69 & 0.255 & 0.111 & 0.670 & 0.881  \\ 
\midrule[1pt]

\textbf{EtriCA} & 2.88 & 49.29 & 22.59 & 41.43 & 0.506 & 0.354 & 0.244 & 0.799       & 8.11 & 39.90 & 9.65 & 25.21 & 0.387 & 0.202 & \textbf{0.359} & \textbf{0.889} \\ 

\textbf{KeEtriCA} & \textbf{2.69} & \textbf{51.85} & \textbf{25.93} & \textbf{44.51} & \textbf{0.535} & \textbf{0.388} & 0.253 & 0.796       & \textbf{7.08} & \textbf{42.31} & \textbf{11.73} & \textbf{27.74} & \textbf{0.419} & \textbf{0.235} & 0.412 & 0.860 \\

\midrule[1pt]
\textbf{Golden} - & - & - & - & - & - & - & 0.048 & 0.906   & - & - & - & - & - & - & 0.286 & 0.950 \\
\bottomrule[2pt]
\end{tabular}
}
\caption{The extension of automatic evaluation on ROC and WP datasets. We extended the automatic evaluation to include the comparison of generated stories among KeEtriCA, EtriCA, ChatGPT, and ChatGLM on the ROC and WP datasets.}
\label{tab:auto-extension}
\end{table*}

\begin{table}[tb]
\centering
\resizebox{0.80\linewidth}{!}{
\begin{tabular}{l|cccc|cccc}
\toprule[1pt]
\multirow{2}{*}{\textbf{Score (1 to 5)}} & \multicolumn{4}{c|}{\textbf{ROC Stories}} & \multicolumn{4}{c}{\textbf{Wringting Prompts}} \\
\cline{2-5} \cline{6-9} 
& \textbf{Fluency} & \textbf{Coherence}  & \textbf{Relevance} & \textbf{Overall}   & \textbf{Fluency} & \textbf{Coherence}  & \textbf{Relevance} & \textbf{Overall} \\
\midrule
\textbf{ChatGLM}  & 4.36 & 4.18 & 4.54 & 4.36      & 3.94 & 3.41 & 3.70 & 3.66   \\ 
\textbf{ChatGPT} & \textbf{4.45} & \textbf{4.44} & \textbf{4.69} & \textbf{4.55}      & \textbf{4.01} & \textbf{4.00} & \textbf{4.49} & \textbf{4.21}   \\ 
\midrule
\textbf{EtriCA} & 3.96 & 3.34 & 3.63 & 3.63      & 3.32 & 2.44 & 2.22 & 2.58 \\ 
\textbf{KeEtriCA} & 4.17 & 3.71 & 4.00 & 3.96      & 3.77 & 2.83 & 2.88 &  3.10   \\ 
\midrule
\textbf{Golden} & 4.40 & 4.40 & 4.67 & 4.49     & 3.98 & 3.85 & 4.14 & 3.96  \\
\bottomrule[1pt]
\end{tabular}
}
\caption{Results of Human Evaluation of KeEtriCA. We calculate Fleiss' Kappa $\kappa$ for each metric. The majority of results indicate a moderate level of agreement ($\kappa \in (0.4, 0.6]$). It should be noted that the Writing Prompts dataset, which consists of story generation prompts gathered from conversations on Reddit, may have lower quality compared to traditional story datasets. We hypothesise that this discrepancy is the reason behind the significantly better performance of \textbf{ChatGPT} compared to the \textbf{Golden} model.}
\label{tab:human_eval_keetrica}
\end{table}

\begin{table*}[ht]
\centering
\resizebox{0.95\linewidth}{!}{
\begin{tabular}{l|ll}
\toprule[2pt]
 \multirow{2}{*}{\textbf{Input}} & \textbf{Leading Context} & [MALE] had lost his dog over a month ago .\\
& \textbf{Event Sequence} & missed dog $\rightarrow$ notices something $\rightarrow$ sees dog $\rightarrow$ turns out be \\
\midrule[1pt]
\textbf{ChatGLM} & \multicolumn{2}{p{0.9\textwidth}}{[MALE] had lost his dog over a month ago, and he had missed him dearly. However, as he was walking his new neighbor's dog, he \textbf{noticed something} that attracts his attention. He \textbf{saw} the \textbf{dog} on the other side of the fence, and he was excited to see him. However, it \textbf{turns out} that the dog \textbf{was} not his dog, but a neighbor's dog that had been adopted out to a local shelter. [MALE] felt a sense of loss and sadness, but he also realized that he had made a new friend in the process.} \\
\midrule
\textbf{ChatGPT} & \multicolumn{2}{p{0.9\textwidth}}{He \textbf{missed} his \textbf{dog} dearly, longing for its company every day. One afternoon, while walking down the street, he \textbf{notices something} familiar out of the corner of his eye. As he approaches, his heart skips a beat - it's his beloved dog! It \textbf{turns out} that the dog he sees \textbf{is} indeed his long-lost companion, happily wagging its tail.}\\
\midrule
\textbf{EtriCA} & \multicolumn{2}{p{0.9\textwidth}}{He \textbf{missed} his \textbf{dog} badly. He \textbf{notices something} strange on the curb. He \textbf{sees} the \textbf{dog} outside. It \textbf{turns out} to \textbf{be} a stray dog.}  \\
\midrule
\textbf{KeEtriCA} & \multicolumn{2}{p{0.9\textwidth}}{He \textbf{missed} his \textbf{dog} most of the time. One day while walking, he \textbf{notices something} strange in the grass. He \textbf{sees} a \textbf{dog} in the grass, holding its face. However, the dog \textbf{turns out} to \textbf{be} a stray dog and [MALE] decided to leave.} \\
\midrule
\textbf{Golden} & \multicolumn{2}{p{0.9\textwidth}}{He \textbf{missed} his \textbf{dog} very much. One evening while mowing the lawn he \textbf{notices something}. He \textbf{sees} a \textbf{dog} in the street that looked like his lost dog. It \textbf{turns out} to \textbf{be} his lost dog who had returned home.} \\
\bottomrule[2pt]
\end{tabular}
}
\caption{This is the extension of the case study. \textit{[MALE]}, \textit{[FEMALE]}, and \textit{[NEUTRAL]} are the specital tokens to replace names in story. The highlighted bold words denote the events corresponding to the given event sequence.}
\label{tab:case_study_extension}
\end{table*}

\subsection{Extension of KeEtriCA} 
By implementing a post-training framework on BookCorpus, KeEtriCA, which relies on an extensive training process, enhances its capability to integrate leading contexts and event sequences for story generation. In this section, we aim to compare the performance improvement achieved by KeEtriCA to its predecessor EtriCA. Furthermore, we include two state-of-the-art Pre-trained Language Models (PLMs), namely ChatGLM and ChatGPT, in this comparative analysis.

ChatGLM and ChatGPT are highly prominent language models widely employed for generating conversational responses. ChatGLM, also known as Chat-based Generative Language Model, is specifically designed to generate responses in chat-based conversations. It is built upon the Generative Language Modeling (GLM) framework and has undergone fine-tuning on an extensive corpus of chat-based data. By leveraging the capabilities of deep neural networks, ChatGLM generates coherent and contextually relevant responses when presented with an input prompt.

On the other hand, ChatGPT, which stands for Chat-based GPT (Generative Pre-trained Transformer), is built upon the GPT architecture—a state-of-the-art language model renowned for its ability to generate high-quality text. ChatGPT is fine-tuned specifically for chat-based conversations, enabling it to produce responses that are not only more engaging but also more context-aware. It incorporates the Transformer model, which employs self-attention mechanisms to effectively capture dependencies and relationships between words within a text sequence. Both ChatGLM and ChatGPT greatly benefit from pre-training on large-scale text corpora, enabling them to acquire a comprehensive understanding of the statistical patterns and linguistic structures inherent in natural language.

It is important to note that the PLMs, namely ChatGLM and ChatGPT, possess significantly larger parameter scales, exceeding 130 billion parameters, in contrast to our models. For instance, EtriCA and KeEtriCA have only 1 billion parameters. We subject both ChatGLM and ChatGPT to zero-shot settings as they are too large to fit on our datasets, requiring a high volume of computing resources. Therefore, in this study we exclude them as the baseline models, instead we provide their results as a reference. However, our method can be also applied to these large PLMS, and this expansion will be left to the future work (ChatGPT is also not open-source up to now).

\autoref{tab:auto-extension} presents the extended experimental results of automatic evaluation, comparing the performance of ChatGLM, ChatGPT, EtriCA, and KeEtriCA. On the other hand,  \autoref{tab:human_eval_keetrica} displays the results of the human evaluation, where three evaluators were presented with 100 samples of stories generated by each model, along with their corresponding input pairs. Evaluators were instructed to rate the stories on a Likert scale ranging from 1 to 5, enabling them to provide a subjective assessment based on their expertise and judgment. The evaluation metrics encompass the previously mentioned criteria of Fluency, Coherence, and Relevance, supplemented by an Overall score.

Both \autoref{tab:auto-extension} and \autoref{tab:human_eval_keetrica} illustrate the substantial improvement achieved by KeEtriCA through post-training on a large corpus, surpassing EtriCA in both automatic and human evaluations. On the short story dataset, ROC Stories, the average increase in automatic evaluation metrics is $4.9\%$, while the human evaluation metrics show a $9.1\%$ improvement.\footnote{The average increase is calculated based on the average improvement across all metrics.} On the long story dataset, Writing Prompts, the automatic and human evaluation exhibit improvements of $7.6\%$ and $20.2\%$ respectively. Furthermore, KeEtriCA outperforms ChatGLM and ChatGPT on most metrics as shown in \autoref{tab:auto-extension}. However, in human evaluation, KeEtriCA is notably inferior to ChatGPT and ChatGLM, suggesting that KeEtriCA excels at adhering to given leading contexts and event sequences in generating stories, but there is still room for improvement in the quality of the generated stories. 

This hypothesis is further supported by the examples presented in Table \ref{tab:case_study_extension}, where both ChatGLM and ChatGPT generate longer and more informative stories, albeit with occasional replacement of events by similar words. For example, "missed dog" is replaced by "missed him" in the story generated by ChatGLM, and "sees dog" is missing in the story generated by ChatGPT. These discrepancies contribute to the lower metrics observed for ChatGPT and ChatGLM in Table \ref{tab:auto-extension}. In addition, KeEtriCA demonstrates improved story quality over that of EtriCA by utilising more similar expressions to those found in human-written stories. For instance, "One day while walking" is replaced by "One evening while mowing the lawn," and "sees a dog in the grass" is replaced by "sees a dog in the street." These similar expressions make the story generated by KeEtriCA read better overall than that of EtriCA. Future work could explore the implementation of ChatGPT or ChatGLM as base language models, as this holds promise for enhancing the ability of neural networks to generate compelling stories.

\section{Conclusion}
We introduce a novel controllable story generation task that involves the conditioning of leading contexts and event sequences. In addition, we present two newly annotated datasets that include extra events and propose a set of automatic metrics to evaluate the coherence and fluency of the generated stories. To address this task, we propose EtriCA, a novel generation model that effectively leverages context and event features through a cross-attention based contextualising network. Furthermore, we enhance the capabilities of EtriCA by employing a post-training framework, which involves additional training using a diverse range of story examples from the BookCorpus dataset. Through extensive experiments and thorough analysis, we demonstrate that both EtriCA and KeEtriCA outperform competitive baselines in terms of fluency, coherence, and relevance in story generation.

\section*{Acknowledgements}
Chen Tang is supported by the China Scholarship Council (CSC) for his doctoral study (File No.202006120039). Tyler Loakman is supported by the Centre for Doctoral Training in Speech and Language Technologies (SLT) and their Applications funded by UK Research and Innovation [Grant number EP/S023062/1]. We also gratefully acknowledge the anonymous reviewers for their insightful comments.

\appendix

\section{Appendix}
\subsection{Details of Human Evaluation}
\begin{figure*}[ht]
\centering
\includegraphics[width=\linewidth]{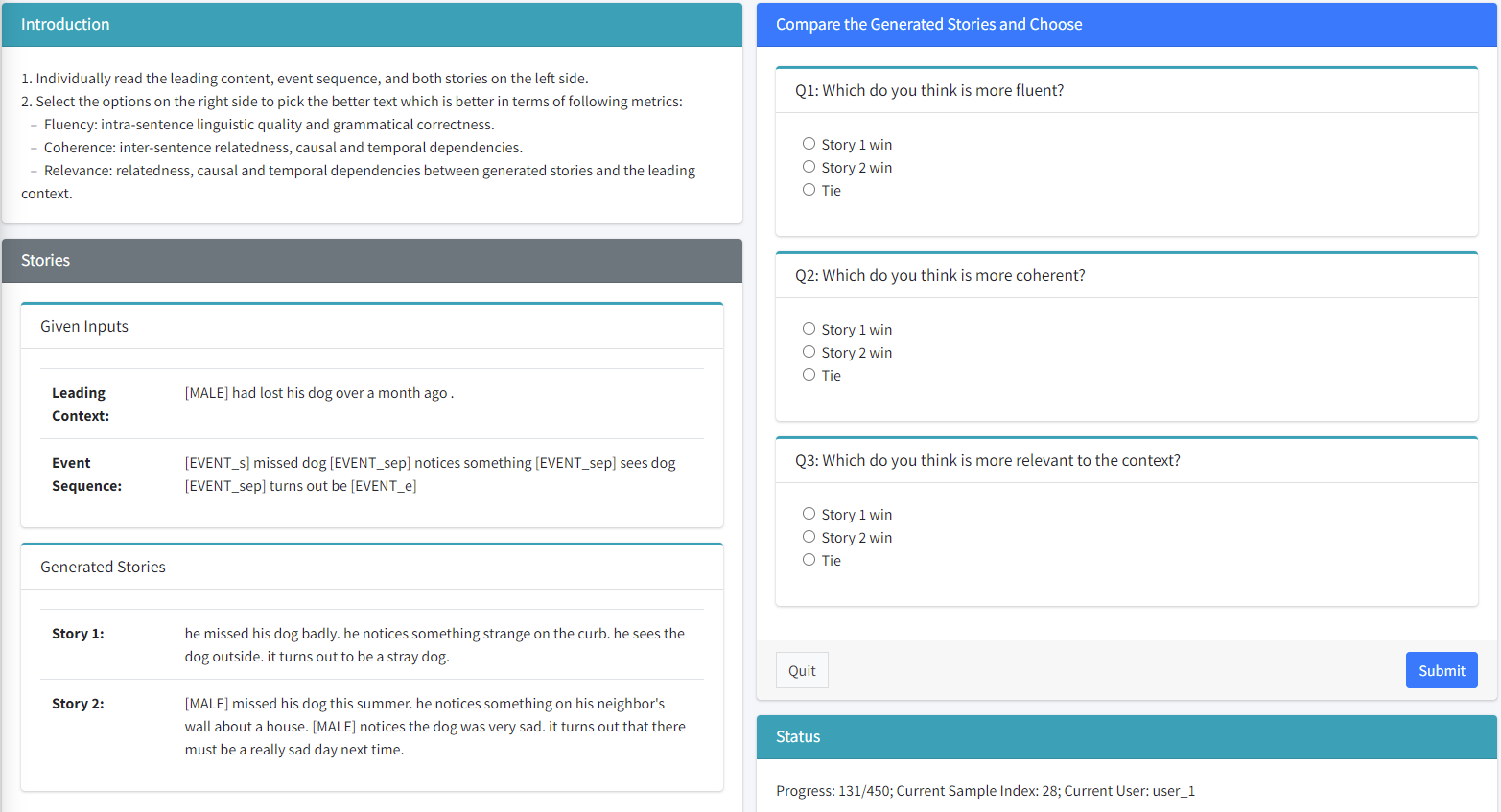}
\caption{Screenshot of our evaluation interface. Stories used in the evaluation are randomly selected from the \textbf{ROC} dataset. Annotators are tasked with making selections for each question presented in the right column. To facilitate precise and accurate annotation, our system permits annotators to perform direct comparisons between various generated stories corresponding to a given input. An automatic recording of the response occurs once all three questions have been answered, and the "submit" button is pressed.}
\label{fig:survey}
\end{figure*}

We have designed an evaluation system to facilitate various tasks, encompassing the collection of evaluation annotations, anonymisation of story pairs for annotation, equitable shuffling of examples, and simplifying the comparison process. As depicted in Figure~\ref{fig:survey}, we provide an overview of our annotation procedure.

Evaluators are required to adhere to the annotation standards outlined in the top-left corner. Recognising the potential variations in individual biases, we inform each annotator of the specific standards established for this task:
\begin{itemize}
\item  \textbf{Fluency}: This dimension evaluates the quality of the generated text, considering aspects such as grammatical errors, spelling errors, unnatural repetitions, and overall language quality. It follows the hierarchy: grammatical errors  $\geq$ spelling errors  $\geq$ unnatural repetitions  $\geq$ language quality.

\item  \textbf{Coherence}: Coherence centers on the logical cohesion between sentences within the generated stories. Annotators are tasked with identifying incoherent segments and assessing the number of word edits required to render the story coherent. Fewer edits needed indicate a more coherent story.

\item  \textbf{Relevance}: Relevance pertains to the alignment between the generated sentences and the provided leading context. However, we acknowledge the subjective nature of determining whether a story is "interesting" or relevant. Therefore, evaluators are instructed to gauge the level of irrelevance in a story by counting the number of generated sentences that conflict with the leading context.
\end{itemize}

\subsection{Our other works}
Our research group has actively contributed to the field of Natural Language Generation (NLG) through various scholarly endeavors. We present a comprehensive categorization of our contributions as follows: Dialogue Generation~\cite{tang-etal-2023-enhancing,yang2023improving}, data-to-text~\cite{yang2023effective}, text summarisation~\cite{tang2023improving,goldsack2023enhancing} and tongue twister generation~\cite{loakman-etal-2023-twistlist}. We believe these works will aid you in navigating our contributions in the field of NLG.

\printcredits

\bibliographystyle{cas-model2-names}

\bibliography{cas-refs}

\begin{thebibliography}{58}
\expandafter\ifx\csname natexlab\endcsname\relax\def\natexlab#1{#1}\fi
\providecommand{\url}[1]{\texttt{#1}}
\providecommand{\href}[2]{#2}
\providecommand{\path}[1]{#1}
\providecommand{\DOIprefix}{doi:}
\providecommand{\ArXivprefix}{arXiv:}
\providecommand{\URLprefix}{URL: }
\providecommand{\Pubmedprefix}{pmid:}
\providecommand{\doi}[1]{\href{http://dx.doi.org/#1}{\path{#1}}}
\providecommand{\Pubmed}[1]{\href{pmid:#1}{\path{#1}}}
\providecommand{\bibinfo}[2]{#2}
\ifx\xfnm\relax \def\xfnm[#1]{\unskip,\space#1}\fi
\bibitem[{Abd~Yusof et~al.(2017)Abd~Yusof, Lin and Guerin}]{abd2017analysing}
\bibinfo{author}{Abd~Yusof, N.F.}, \bibinfo{author}{Lin, C.},
  \bibinfo{author}{Guerin, F.}, \bibinfo{year}{2017}.
\newblock \bibinfo{title}{Analysing the causes of depressed mood from
  depression vulnerable individuals}, in: \bibinfo{booktitle}{Proceedings of
  the International Workshop on Digital Disease Detection using Social Media
  2017 (DDDSM-2017)}, pp. \bibinfo{pages}{9--17}.
\bibitem[{Alhussain and Azmi(2021)}]{alhussain2021automatic}
\bibinfo{author}{Alhussain, A.I.}, \bibinfo{author}{Azmi, A.M.},
  \bibinfo{year}{2021}.
\newblock \bibinfo{title}{Automatic story generation: a survey of approaches}.
\newblock \bibinfo{journal}{ACM Computing Surveys (CSUR)} \bibinfo{volume}{54},
  \bibinfo{pages}{1--38}.
\bibitem[{Bird and Loper(2004)}]{bird-loper-2004-nltk}
\bibinfo{author}{Bird, S.}, \bibinfo{author}{Loper, E.}, \bibinfo{year}{2004}.
\newblock \bibinfo{title}{{NLTK}: The natural language toolkit}, in:
  \bibinfo{booktitle}{Proceedings of the {ACL} Interactive Poster and
  Demonstration Sessions}, \bibinfo{address}{Barcelona, Spain}. pp.
  \bibinfo{pages}{214--217}.
\newblock \URLprefix \url{https://aclanthology.org/P04-3031}.
\bibitem[{Bj{\"o}rne and Salakoski(2018)}]{bjorne-salakoski-2018-biomedical}
\bibinfo{author}{Bj{\"o}rne, J.}, \bibinfo{author}{Salakoski, T.},
  \bibinfo{year}{2018}.
\newblock \bibinfo{title}{Biomedical event extraction using convolutional
  neural networks and dependency parsing}, in: \bibinfo{booktitle}{Proceedings
  of the {B}io{NLP} 2018 workshop}, \bibinfo{address}{Melbourne, Australia}.
  pp. \bibinfo{pages}{98--108}.
\newblock \URLprefix \url{https://aclanthology.org/W18-2311},
  \DOIprefix\doi{10.18653/v1/W18-2311}.
\bibitem[{Chen et~al.(2018)Chen, Du, Bing and Xu}]{chen-etal-2018-hybrid}
\bibinfo{author}{Chen, D.}, \bibinfo{author}{Du, J.}, \bibinfo{author}{Bing,
  L.}, \bibinfo{author}{Xu, R.}, \bibinfo{year}{2018}.
\newblock \bibinfo{title}{Hybrid neural attention for agreement/disagreement
  inference in online debates}, in: \bibinfo{booktitle}{Proceedings of the 2018
  Conference on Empirical Methods in Natural Language Processing},
  \bibinfo{publisher}{Association for Computational Linguistics},
  \bibinfo{address}{Brussels, Belgium}. pp. \bibinfo{pages}{665--670}.
\newblock \URLprefix \url{https://aclanthology.org/D18-1069},
  \DOIprefix\doi{10.18653/v1/D18-1069}.
\bibitem[{Chen et~al.(2021)Chen, Shu, Takamura and
  Nakayama}]{chen-etal-2021-graphplan}
\bibinfo{author}{Chen, H.}, \bibinfo{author}{Shu, R.},
  \bibinfo{author}{Takamura, H.}, \bibinfo{author}{Nakayama, H.},
  \bibinfo{year}{2021}.
\newblock \bibinfo{title}{{G}raph{P}lan: Story generation by planning with
  event graph}, in: \bibinfo{booktitle}{Proceedings of the 14th International
  Conference on Natural Language Generation}, \bibinfo{publisher}{Association
  for Computational Linguistics}, \bibinfo{address}{Aberdeen, Scotland, UK}.
  pp. \bibinfo{pages}{377--386}.
\newblock \URLprefix \url{https://aclanthology.org/2021.inlg-1.42}.
\bibitem[{Clark and Smith(2021)}]{clark-smith-2021-choose}
\bibinfo{author}{Clark, E.}, \bibinfo{author}{Smith, N.A.},
  \bibinfo{year}{2021}.
\newblock \bibinfo{title}{Choose your own adventure: Paired suggestions in
  collaborative writing for evaluating story generation models}, in:
  \bibinfo{booktitle}{Proceedings of the 2021 Conference of the North American
  Chapter of the Association for Computational Linguistics: Human Language
  Technologies}, \bibinfo{address}{Online}. pp. \bibinfo{pages}{3566--3575}.
\newblock \URLprefix \url{https://aclanthology.org/2021.naacl-main.279},
  \DOIprefix\doi{10.18653/v1/2021.naacl-main.279}.
\bibitem[{De~Marneffe and Manning(2008)}]{de2008stanford}
\bibinfo{author}{De~Marneffe, M.C.}, \bibinfo{author}{Manning, C.D.},
  \bibinfo{year}{2008}.
\newblock \bibinfo{title}{Stanford typed dependencies manual}.
\newblock \bibinfo{type}{Technical Report}. Technical report, Stanford
  University.
\bibitem[{Devlin et~al.(2019)Devlin, Chang, Lee and
  Toutanova}]{devlin-etal-2019-bert}
\bibinfo{author}{Devlin, J.}, \bibinfo{author}{Chang, M.W.},
  \bibinfo{author}{Lee, K.}, \bibinfo{author}{Toutanova, K.},
  \bibinfo{year}{2019}.
\newblock \bibinfo{title}{{BERT}: Pre-training of deep bidirectional
  transformers for language understanding}, in: \bibinfo{booktitle}{Proceedings
  of the 2019 Conference of the North {A}merican Chapter of the Association for
  Computational Linguistics: Human Language Technologies, Volume 1 (Long and
  Short Papers)}, \bibinfo{address}{Minneapolis, Minnesota}. pp.
  \bibinfo{pages}{4171--4186}.
\newblock \URLprefix \url{https://aclanthology.org/N19-1423},
  \DOIprefix\doi{10.18653/v1/N19-1423}.
\bibitem[{Fan et~al.(2018)Fan, Lewis and Dauphin}]{fan-etal-2018-hierarchical}
\bibinfo{author}{Fan, A.}, \bibinfo{author}{Lewis, M.},
  \bibinfo{author}{Dauphin, Y.}, \bibinfo{year}{2018}.
\newblock \bibinfo{title}{Hierarchical neural story generation}, in:
  \bibinfo{booktitle}{Proceedings of the 56th Annual Meeting of the Association
  for Computational Linguistics (Volume 1: Long Papers)},
  \bibinfo{publisher}{Association for Computational Linguistics},
  \bibinfo{address}{Melbourne, Australia}. pp. \bibinfo{pages}{889--898}.
\newblock \URLprefix \url{https://aclanthology.org/P18-1082},
  \DOIprefix\doi{10.18653/v1/P18-1082}.
\bibitem[{Fleiss(1971)}]{fleiss1971measuring}
\bibinfo{author}{Fleiss, J.L.}, \bibinfo{year}{1971}.
\newblock \bibinfo{title}{Measuring nominal scale agreement among many raters.}
\newblock \bibinfo{journal}{Psychological bulletin} \bibinfo{volume}{76},
  \bibinfo{pages}{378}.
\bibitem[{Garg et~al.(2019)Garg, Peitz, Nallasamy and
  Paulik}]{garg-etal-2019-jointly}
\bibinfo{author}{Garg, S.}, \bibinfo{author}{Peitz, S.},
  \bibinfo{author}{Nallasamy, U.}, \bibinfo{author}{Paulik, M.},
  \bibinfo{year}{2019}.
\newblock \bibinfo{title}{Jointly learning to align and translate with
  transformer models}, in: \bibinfo{booktitle}{Proceedings of the 2019
  Conference on Empirical Methods in Natural Language Processing and the 9th
  International Joint Conference on Natural Language Processing
  (EMNLP-IJCNLP)}, \bibinfo{address}{Hong Kong, China}. pp.
  \bibinfo{pages}{4453--4462}.
\newblock \URLprefix \url{https://aclanthology.org/D19-1453},
  \DOIprefix\doi{10.18653/v1/D19-1453}.
\bibitem[{Ghazarian et~al.(2021)Ghazarian, Liu, S~M, Weischedel, Galstyan and
  Peng}]{ghazarian-etal-2021-plot}
\bibinfo{author}{Ghazarian, S.}, \bibinfo{author}{Liu, Z.},
  \bibinfo{author}{S~M, A.}, \bibinfo{author}{Weischedel, R.},
  \bibinfo{author}{Galstyan, A.}, \bibinfo{author}{Peng, N.},
  \bibinfo{year}{2021}.
\newblock \bibinfo{title}{Plot-guided adversarial example construction for
  evaluating open-domain story generation}, in: \bibinfo{booktitle}{Proceedings
  of the 2021 Conference of the North American Chapter of the Association for
  Computational Linguistics: Human Language Technologies},
  \bibinfo{publisher}{Association for Computational Linguistics},
  \bibinfo{address}{Online}. pp. \bibinfo{pages}{4334--4344}.
\newblock \URLprefix \url{https://aclanthology.org/2021.naacl-main.343},
  \DOIprefix\doi{10.18653/v1/2021.naacl-main.343}.
\bibitem[{Gheini et~al.(2021)Gheini, Ren and May}]{gheini2021cross}
\bibinfo{author}{Gheini, M.}, \bibinfo{author}{Ren, X.}, \bibinfo{author}{May,
  J.}, \bibinfo{year}{2021}.
\newblock \bibinfo{title}{Cross-attention is all you need: Adapting pretrained
  transformers for machine translation}, in: \bibinfo{editor}{Moens, M.},
  \bibinfo{editor}{Huang, X.}, \bibinfo{editor}{Specia, L.},
  \bibinfo{editor}{Yih, S.W.} (Eds.), \bibinfo{booktitle}{Proceedings of the
  2021 Conference on Empirical Methods in Natural Language Processing, {EMNLP}
  2021, Virtual Event / Punta Cana, Dominican Republic, 7-11 November, 2021},
  \bibinfo{publisher}{Association for Computational Linguistics}. pp.
  \bibinfo{pages}{1754--1765}.
\newblock \URLprefix \url{https://doi.org/10.18653/v1/2021.emnlp-main.132},
  \DOIprefix\doi{10.18653/v1/2021.emnlp-main.132}.
\bibitem[{Goldfarb-Tarrant et~al.(2020)Goldfarb-Tarrant, Chakrabarty,
  Weischedel and Peng}]{goldfarb-tarrant-etal-2020-content}
\bibinfo{author}{Goldfarb-Tarrant, S.}, \bibinfo{author}{Chakrabarty, T.},
  \bibinfo{author}{Weischedel, R.}, \bibinfo{author}{Peng, N.},
  \bibinfo{year}{2020}.
\newblock \bibinfo{title}{Content planning for neural story generation with
  aristotelian rescoring}, in: \bibinfo{booktitle}{Proceedings of the 2020
  Conference on Empirical Methods in Natural Language Processing (EMNLP)},
  \bibinfo{publisher}{Association for Computational Linguistics},
  \bibinfo{address}{Online}. pp. \bibinfo{pages}{4319--4338}.
\newblock \URLprefix \url{https://aclanthology.org/2020.emnlp-main.351},
  \DOIprefix\doi{10.18653/v1/2020.emnlp-main.351}.
\bibitem[{Goldsack et~al.(2023)Goldsack, Zhang, Tang, Scarton and
  Lin}]{goldsack2023enhancing}
\bibinfo{author}{Goldsack, T.}, \bibinfo{author}{Zhang, Z.},
  \bibinfo{author}{Tang, C.}, \bibinfo{author}{Scarton, C.},
  \bibinfo{author}{Lin, C.}, \bibinfo{year}{2023}.
\newblock \bibinfo{title}{Enhancing biomedical lay summarisation with external
  knowledge graphs}.
\newblock \bibinfo{journal}{arXiv preprint arXiv:2310.15702} .
\bibitem[{Guan et~al.(2020)Guan, Huang, Zhao, Zhu and
  Huang}]{guan-etal-2020-knowledge}
\bibinfo{author}{Guan, J.}, \bibinfo{author}{Huang, F.}, \bibinfo{author}{Zhao,
  Z.}, \bibinfo{author}{Zhu, X.}, \bibinfo{author}{Huang, M.},
  \bibinfo{year}{2020}.
\newblock \bibinfo{title}{A knowledge-enhanced pretraining model for
  commonsense story generation}.
\newblock \bibinfo{journal}{Transactions of the Association for Computational
  Linguistics} \bibinfo{volume}{8}, \bibinfo{pages}{93--108}.
\newblock \URLprefix \url{https://aclanthology.org/2020.tacl-1.7},
  \DOIprefix\doi{10.1162/tacl_a_00302}.
\bibitem[{Guan et~al.(2021)Guan, Mao, Fan, Liu, Ding and
  Huang}]{guan-etal-2021-long}
\bibinfo{author}{Guan, J.}, \bibinfo{author}{Mao, X.}, \bibinfo{author}{Fan,
  C.}, \bibinfo{author}{Liu, Z.}, \bibinfo{author}{Ding, W.},
  \bibinfo{author}{Huang, M.}, \bibinfo{year}{2021}.
\newblock \bibinfo{title}{Long text generation by modeling sentence-level and
  discourse-level coherence}, in: \bibinfo{booktitle}{Proceedings of the 59th
  Annual Meeting of the Association for Computational Linguistics and the 11th
  International Joint Conference on Natural Language Processing (Volume 1: Long
  Papers)}, \bibinfo{address}{Online}. pp. \bibinfo{pages}{6379--6393}.
\newblock \URLprefix \url{https://aclanthology.org/2021.acl-long.499},
  \DOIprefix\doi{10.18653/v1/2021.acl-long.499}.
\bibitem[{He et~al.(2020)He, Tran, Haffari, Chang, Lin, Bui, Dernoncourt and
  Dam}]{he-etal-2020-scene}
\bibinfo{author}{He, X.}, \bibinfo{author}{Tran, Q.H.},
  \bibinfo{author}{Haffari, G.}, \bibinfo{author}{Chang, W.},
  \bibinfo{author}{Lin, Z.}, \bibinfo{author}{Bui, T.},
  \bibinfo{author}{Dernoncourt, F.}, \bibinfo{author}{Dam, N.},
  \bibinfo{year}{2020}.
\newblock \bibinfo{title}{Scene graph modification based on natural language
  commands}, in: \bibinfo{booktitle}{Findings of the Association for
  Computational Linguistics: EMNLP 2020}, \bibinfo{publisher}{Association for
  Computational Linguistics}, \bibinfo{address}{Online}. pp.
  \bibinfo{pages}{972--990}.
\newblock \URLprefix \url{https://aclanthology.org/2020.findings-emnlp.87},
  \DOIprefix\doi{10.18653/v1/2020.findings-emnlp.87}.
\bibitem[{Holtzman et~al.(2019)Holtzman, Buys, Du, Forbes and
  Choi}]{holtzman2019curious}
\bibinfo{author}{Holtzman, A.}, \bibinfo{author}{Buys, J.},
  \bibinfo{author}{Du, L.}, \bibinfo{author}{Forbes, M.},
  \bibinfo{author}{Choi, Y.}, \bibinfo{year}{2019}.
\newblock \bibinfo{title}{The curious case of neural text degeneration}.
\newblock \bibinfo{journal}{arXiv preprint arXiv:1904.09751} .
\bibitem[{Huang et~al.(2022)Huang, Tang, Loakman, Guerin and
  Lin}]{huang-etal-2022-improving}
\bibinfo{author}{Huang, H.}, \bibinfo{author}{Tang, C.},
  \bibinfo{author}{Loakman, T.}, \bibinfo{author}{Guerin, F.},
  \bibinfo{author}{Lin, C.}, \bibinfo{year}{2022}.
\newblock \bibinfo{title}{Improving {C}hinese story generation via awareness of
  syntactic dependencies and semantics}, in: \bibinfo{booktitle}{Proceedings of
  the 2nd Conference of the Asia-Pacific Chapter of the Association for
  Computational Linguistics and the 12th International Joint Conference on
  Natural Language Processing (Volume 2: Short Papers)},
  \bibinfo{publisher}{Association for Computational Linguistics},
  \bibinfo{address}{Online only}. pp. \bibinfo{pages}{178--185}.
\newblock \URLprefix \url{https://aclanthology.org/2022.aacl-short.23}.
\bibitem[{Huang and Huang(2013)}]{huang-huang-2013-optimized}
\bibinfo{author}{Huang, L.}, \bibinfo{author}{Huang, L.}, \bibinfo{year}{2013}.
\newblock \bibinfo{title}{Optimized event storyline generation based on
  mixture-event-aspect model}, in: \bibinfo{booktitle}{Proceedings of the 2013
  Conference on Empirical Methods in Natural Language Processing},
  \bibinfo{address}{Seattle, Washington, USA}. pp. \bibinfo{pages}{726--735}.
\newblock \URLprefix \url{https://aclanthology.org/D13-1068}.
\bibitem[{Huang et~al.(2018)Huang, Ji, Cho, Dagan, Riedel and
  Voss}]{huang-etal-2018-zero}
\bibinfo{author}{Huang, L.}, \bibinfo{author}{Ji, H.}, \bibinfo{author}{Cho,
  K.}, \bibinfo{author}{Dagan, I.}, \bibinfo{author}{Riedel, S.},
  \bibinfo{author}{Voss, C.}, \bibinfo{year}{2018}.
\newblock \bibinfo{title}{Zero-shot transfer learning for event extraction},
  in: \bibinfo{booktitle}{Proceedings of the 56th Annual Meeting of the
  Association for Computational Linguistics (Volume 1: Long Papers)},
  \bibinfo{address}{Melbourne, Australia}. pp. \bibinfo{pages}{2160--2170}.
\newblock \URLprefix \url{https://aclanthology.org/P18-1201},
  \DOIprefix\doi{10.18653/v1/P18-1201}.
\bibitem[{Jhamtani and
  Berg-Kirkpatrick(2020)}]{jhamtani-berg-kirkpatrick-2020-narrative}
\bibinfo{author}{Jhamtani, H.}, \bibinfo{author}{Berg-Kirkpatrick, T.},
  \bibinfo{year}{2020}.
\newblock \bibinfo{title}{Narrative text generation with a latent discrete
  plan}, in: \bibinfo{booktitle}{Findings of the Association for Computational
  Linguistics: EMNLP 2020}, \bibinfo{address}{Online}. pp.
  \bibinfo{pages}{3637--3650}.
\newblock \URLprefix \url{https://aclanthology.org/2020.findings-emnlp.325},
  \DOIprefix\doi{10.18653/v1/2020.findings-emnlp.325}.
\bibitem[{Kingma and Ba(2014)}]{kingma2014adam}
\bibinfo{author}{Kingma, D.P.}, \bibinfo{author}{Ba, J.}, \bibinfo{year}{2014}.
\newblock \bibinfo{title}{Adam: A method for stochastic optimization}.
\newblock \bibinfo{journal}{arXiv preprint arXiv:1412.6980} .
\bibitem[{Kong et~al.(2021)Kong, Huang, Tung, Guan and
  Huang}]{kong2021stylized}
\bibinfo{author}{Kong, X.}, \bibinfo{author}{Huang, J.}, \bibinfo{author}{Tung,
  Z.}, \bibinfo{author}{Guan, J.}, \bibinfo{author}{Huang, M.},
  \bibinfo{year}{2021}.
\newblock \bibinfo{title}{Stylized story generation with style-guided
  planning}.
\newblock \bibinfo{journal}{arXiv preprint arXiv:2105.08625} .
\bibitem[{Kybartas and Bidarra(2016)}]{kybartas2016survey}
\bibinfo{author}{Kybartas, B.}, \bibinfo{author}{Bidarra, R.},
  \bibinfo{year}{2016}.
\newblock \bibinfo{title}{A survey on story generation techniques for authoring
  computational narratives}.
\newblock \bibinfo{journal}{IEEE Transactions on Computational Intelligence and
  AI in Games} \bibinfo{volume}{9}, \bibinfo{pages}{239--253}.
\bibitem[{Lewis et~al.(2020)Lewis, Liu, Goyal, Ghazvininejad, Mohamed, Levy,
  Stoyanov and Zettlemoyer}]{lewis-etal-2020-bart}
\bibinfo{author}{Lewis, M.}, \bibinfo{author}{Liu, Y.}, \bibinfo{author}{Goyal,
  N.}, \bibinfo{author}{Ghazvininejad, M.}, \bibinfo{author}{Mohamed, A.},
  \bibinfo{author}{Levy, O.}, \bibinfo{author}{Stoyanov, V.},
  \bibinfo{author}{Zettlemoyer, L.}, \bibinfo{year}{2020}.
\newblock \bibinfo{title}{{BART}: Denoising sequence-to-sequence pre-training
  for natural language generation, translation, and comprehension}, in:
  \bibinfo{booktitle}{Proceedings of the 58th Annual Meeting of the Association
  for Computational Linguistics}, \bibinfo{publisher}{Association for
  Computational Linguistics}, \bibinfo{address}{Online}. pp.
  \bibinfo{pages}{7871--7880}.
\newblock \URLprefix \url{https://aclanthology.org/2020.acl-main.703},
  \DOIprefix\doi{10.18653/v1/2020.acl-main.703}.
\bibitem[{Li et~al.(2016)Li, Galley, Brockett, Gao and
  Dolan}]{li-etal-2016-diversity}
\bibinfo{author}{Li, J.}, \bibinfo{author}{Galley, M.},
  \bibinfo{author}{Brockett, C.}, \bibinfo{author}{Gao, J.},
  \bibinfo{author}{Dolan, B.}, \bibinfo{year}{2016}.
\newblock \bibinfo{title}{A diversity-promoting objective function for neural
  conversation models}, in: \bibinfo{booktitle}{Proceedings of the 2016
  Conference of the North {A}merican Chapter of the Association for
  Computational Linguistics: Human Language Technologies},
  \bibinfo{address}{San Diego, California}. pp. \bibinfo{pages}{110--119}.
\newblock \URLprefix \url{https://aclanthology.org/N16-1014},
  \DOIprefix\doi{10.18653/v1/N16-1014}.
\bibitem[{Lin(2004)}]{lin-2004-rouge}
\bibinfo{author}{Lin, C.Y.}, \bibinfo{year}{2004}.
\newblock \bibinfo{title}{{ROUGE}: A package for automatic evaluation of
  summaries}, in: \bibinfo{booktitle}{Text Summarization Branches Out},
  \bibinfo{address}{Barcelona, Spain}. pp. \bibinfo{pages}{74--81}.
\newblock \URLprefix \url{https://aclanthology.org/W04-1013}.
\bibitem[{Loakman et~al.(2023)Loakman, Tang and
  Lin}]{loakman-etal-2023-twistlist}
\bibinfo{author}{Loakman, T.}, \bibinfo{author}{Tang, C.},
  \bibinfo{author}{Lin, C.}, \bibinfo{year}{2023}.
\newblock \bibinfo{title}{{T}wist{L}ist: Resources and baselines for tongue
  twister generation}, in: \bibinfo{booktitle}{Proceedings of the 61st Annual
  Meeting of the Association for Computational Linguistics (Volume 2: Short
  Papers)}, \bibinfo{publisher}{Association for Computational Linguistics},
  \bibinfo{address}{Toronto, Canada}. pp. \bibinfo{pages}{579--589}.
\newblock \URLprefix \url{https://aclanthology.org/2023.acl-short.51},
  \DOIprefix\doi{10.18653/v1/2023.acl-short.51}.
\bibitem[{McIntyre and Lapata(2009)}]{mcintyre-lapata-2009-learning}
\bibinfo{author}{McIntyre, N.}, \bibinfo{author}{Lapata, M.},
  \bibinfo{year}{2009}.
\newblock \bibinfo{title}{Learning to tell tales: A data-driven approach to
  story generation}, in: \bibinfo{booktitle}{Proceedings of the Joint
  Conference of the 47th Annual Meeting of the {ACL} and the 4th International
  Joint Conference on Natural Language Processing of the {AFNLP}},
  \bibinfo{address}{Suntec, Singapore}. pp. \bibinfo{pages}{217--225}.
\newblock \URLprefix \url{https://aclanthology.org/P09-1025}.
\bibitem[{McIntyre and Lapata(2010)}]{mcintyre-lapata-2010-plot}
\bibinfo{author}{McIntyre, N.}, \bibinfo{author}{Lapata, M.},
  \bibinfo{year}{2010}.
\newblock \bibinfo{title}{Plot induction and evolutionary search for story
  generation}, in: \bibinfo{booktitle}{Proceedings of the 48th Annual Meeting
  of the Association for Computational Linguistics}, \bibinfo{address}{Uppsala,
  Sweden}. pp. \bibinfo{pages}{1562--1572}.
\newblock \URLprefix \url{https://aclanthology.org/P10-1158}.
\bibitem[{Mostafazadeh et~al.(2016)Mostafazadeh, Chambers, He, Parikh, Batra,
  Vanderwende, Kohli and Allen}]{mostafazadeh-etal-2016-corpus}
\bibinfo{author}{Mostafazadeh, N.}, \bibinfo{author}{Chambers, N.},
  \bibinfo{author}{He, X.}, \bibinfo{author}{Parikh, D.},
  \bibinfo{author}{Batra, D.}, \bibinfo{author}{Vanderwende, L.},
  \bibinfo{author}{Kohli, P.}, \bibinfo{author}{Allen, J.},
  \bibinfo{year}{2016}.
\newblock \bibinfo{title}{A corpus and cloze evaluation for deeper
  understanding of commonsense stories}, in: \bibinfo{booktitle}{Proceedings of
  the 2016 Conference of the North {A}merican Chapter of the Association for
  Computational Linguistics: Human Language Technologies},
  \bibinfo{address}{San Diego, California}. pp. \bibinfo{pages}{839--849}.
\newblock \URLprefix \url{https://aclanthology.org/N16-1098},
  \DOIprefix\doi{10.18653/v1/N16-1098}.
\bibitem[{Papineni et~al.(2002)Papineni, Roukos, Ward and
  Zhu}]{papineni-etal-2002-bleu}
\bibinfo{author}{Papineni, K.}, \bibinfo{author}{Roukos, S.},
  \bibinfo{author}{Ward, T.}, \bibinfo{author}{Zhu, W.J.},
  \bibinfo{year}{2002}.
\newblock \bibinfo{title}{{B}leu: a method for automatic evaluation of machine
  translation}, in: \bibinfo{booktitle}{Proceedings of the 40th Annual Meeting
  of the Association for Computational Linguistics},
  \bibinfo{address}{Philadelphia, Pennsylvania, USA}. pp.
  \bibinfo{pages}{311--318}.
\newblock \URLprefix \url{https://aclanthology.org/P02-1040},
  \DOIprefix\doi{10.3115/1073083.1073135}.
\bibitem[{Peng and Roth(2016)}]{peng-roth-2016-two}
\bibinfo{author}{Peng, H.}, \bibinfo{author}{Roth, D.}, \bibinfo{year}{2016}.
\newblock \bibinfo{title}{Two discourse driven language models for semantics},
  in: \bibinfo{booktitle}{Proceedings of the 54th Annual Meeting of the
  Association for Computational Linguistics (Volume 1: Long Papers)},
  \bibinfo{address}{Berlin, Germany}. pp. \bibinfo{pages}{290--300}.
\newblock \URLprefix \url{https://aclanthology.org/P16-1028},
  \DOIprefix\doi{10.18653/v1/P16-1028}.
\bibitem[{Peng et~al.(2021)Peng, Yin, Rong, Lin, Zhou and
  Xiong}]{peng2021named}
\bibinfo{author}{Peng, K.}, \bibinfo{author}{Yin, C.}, \bibinfo{author}{Rong,
  W.}, \bibinfo{author}{Lin, C.}, \bibinfo{author}{Zhou, D.},
  \bibinfo{author}{Xiong, Z.}, \bibinfo{year}{2021}.
\newblock \bibinfo{title}{Named entity aware transfer learning for biomedical
  factoid question answering}.
\newblock \bibinfo{journal}{IEEE/ACM Transactions on Computational Biology and
  Bioinformatics} \bibinfo{volume}{19}, \bibinfo{pages}{2365--2376}.
\bibitem[{Radford et~al.(2019)Radford, Wu, Child, Luan, Amodei, Sutskever
  et~al.}]{radford2019language}
\bibinfo{author}{Radford, A.}, \bibinfo{author}{Wu, J.},
  \bibinfo{author}{Child, R.}, \bibinfo{author}{Luan, D.},
  \bibinfo{author}{Amodei, D.}, \bibinfo{author}{Sutskever, I.}, et~al.,
  \bibinfo{year}{2019}.
\newblock \bibinfo{title}{Language models are unsupervised multitask learners}.
\newblock \bibinfo{journal}{OpenAI blog} \bibinfo{volume}{1},
  \bibinfo{pages}{9}.
\bibitem[{Rashkin et~al.(2020)Rashkin, Celikyilmaz, Choi and
  Gao}]{rashkin-etal-2020-plotmachines}
\bibinfo{author}{Rashkin, H.}, \bibinfo{author}{Celikyilmaz, A.},
  \bibinfo{author}{Choi, Y.}, \bibinfo{author}{Gao, J.}, \bibinfo{year}{2020}.
\newblock \bibinfo{title}{{P}lot{M}achines: Outline-conditioned generation with
  dynamic plot state tracking}, in: \bibinfo{booktitle}{Proceedings of the 2020
  Conference on Empirical Methods in Natural Language Processing (EMNLP)},
  \bibinfo{publisher}{Association for Computational Linguistics},
  \bibinfo{address}{Online}. pp. \bibinfo{pages}{4274--4295}.
\newblock \URLprefix \url{https://aclanthology.org/2020.emnlp-main.349},
  \DOIprefix\doi{10.18653/v1/2020.emnlp-main.349}.
\bibitem[{Reimers and Gurevych(2019)}]{reimers-gurevych-2019-sentence}
\bibinfo{author}{Reimers, N.}, \bibinfo{author}{Gurevych, I.},
  \bibinfo{year}{2019}.
\newblock \bibinfo{title}{Sentence-{BERT}: Sentence embeddings using {S}iamese
  {BERT}-networks}, in: \bibinfo{booktitle}{Proceedings of the 2019 Conference
  on Empirical Methods in Natural Language Processing and the 9th International
  Joint Conference on Natural Language Processing (EMNLP-IJCNLP)},
  \bibinfo{publisher}{Association for Computational Linguistics},
  \bibinfo{address}{Hong Kong, China}. pp. \bibinfo{pages}{3982--3992}.
\newblock \URLprefix \url{https://aclanthology.org/D19-1410},
  \DOIprefix\doi{10.18653/v1/D19-1410}.
\bibitem[{Rusu et~al.(2014)Rusu, Hodson and
  Kimball}]{rusu-etal-2014-unsupervised}
\bibinfo{author}{Rusu, D.}, \bibinfo{author}{Hodson, J.},
  \bibinfo{author}{Kimball, A.}, \bibinfo{year}{2014}.
\newblock \bibinfo{title}{Unsupervised techniques for extracting and clustering
  complex events in news}, in: \bibinfo{booktitle}{Proceedings of the Second
  Workshop on {EVENTS}: Definition, Detection, Coreference, and
  Representation}, \bibinfo{address}{Baltimore, Maryland, USA}. pp.
  \bibinfo{pages}{26--34}.
\newblock \URLprefix \url{https://aclanthology.org/W14-2905},
  \DOIprefix\doi{10.3115/v1/W14-2905}.
\bibitem[{Shao et~al.(2019)Shao, Huang, Wen, Xu and Zhu}]{shao-etal-2019-long}
\bibinfo{author}{Shao, Z.}, \bibinfo{author}{Huang, M.}, \bibinfo{author}{Wen,
  J.}, \bibinfo{author}{Xu, W.}, \bibinfo{author}{Zhu, X.},
  \bibinfo{year}{2019}.
\newblock \bibinfo{title}{Long and diverse text generation with planning-based
  hierarchical variational model}, in: \bibinfo{booktitle}{Proceedings of the
  2019 Conference on Empirical Methods in Natural Language Processing and the
  9th International Joint Conference on Natural Language Processing
  (EMNLP-IJCNLP)}, \bibinfo{address}{Hong Kong, China}. pp.
  \bibinfo{pages}{3257--3268}.
\newblock \URLprefix \url{https://aclanthology.org/D19-1321},
  \DOIprefix\doi{10.18653/v1/D19-1321}.
\bibitem[{Tang et~al.(2022a)Tang, Guerin, Li and Lin}]{tang2022recent}
\bibinfo{author}{Tang, C.}, \bibinfo{author}{Guerin, F.}, \bibinfo{author}{Li,
  Y.}, \bibinfo{author}{Lin, C.}, \bibinfo{year}{2022}a.
\newblock \bibinfo{title}{Recent advances in neural text generation: A
  task-agnostic survey}.
\newblock \bibinfo{journal}{arXiv preprint arXiv:2203.03047} .
\bibitem[{Tang et~al.(2022b)Tang, Lin, Huang, Guerin and
  Zhang}]{tang-etal-2022-EtriCA}
\bibinfo{author}{Tang, C.}, \bibinfo{author}{Lin, C.}, \bibinfo{author}{Huang,
  H.}, \bibinfo{author}{Guerin, F.}, \bibinfo{author}{Zhang, Z.},
  \bibinfo{year}{2022}b.
\newblock \bibinfo{title}{{E}tri{CA}: Event-triggered context-aware story
  generation augmented by cross attention}, in: \bibinfo{booktitle}{Findings of
  the Association for Computational Linguistics: EMNLP 2022},
  \bibinfo{publisher}{Association for Computational Linguistics},
  \bibinfo{address}{Abu Dhabi, United Arab Emirates}. pp.
  \bibinfo{pages}{5504--5518}.
\newblock \URLprefix \url{https://aclanthology.org/2022.findings-emnlp.403}.
\bibitem[{Tang et~al.(2023a)Tang, Wang, Goldsack and Lin}]{tang2023improving}
\bibinfo{author}{Tang, C.}, \bibinfo{author}{Wang, S.},
  \bibinfo{author}{Goldsack, T.}, \bibinfo{author}{Lin, C.},
  \bibinfo{year}{2023}a.
\newblock \bibinfo{title}{Improving biomedical abstractive summarisation with
  knowledge aggregation from citation papers}.
\newblock \bibinfo{journal}{arXiv preprint arXiv:2310.15684} .
\bibitem[{Tang et~al.(2023b)Tang, Zhang, Loakman, Lin and
  Guerin}]{tang-etal-2023-enhancing}
\bibinfo{author}{Tang, C.}, \bibinfo{author}{Zhang, H.},
  \bibinfo{author}{Loakman, T.}, \bibinfo{author}{Lin, C.},
  \bibinfo{author}{Guerin, F.}, \bibinfo{year}{2023}b.
\newblock \bibinfo{title}{Enhancing dialogue generation via dynamic graph
  knowledge aggregation}, in: \bibinfo{booktitle}{Proceedings of the 61st
  Annual Meeting of the Association for Computational Linguistics (Volume 1:
  Long Papers)}, \bibinfo{publisher}{Association for Computational
  Linguistics}, \bibinfo{address}{Toronto, Canada}. pp.
  \bibinfo{pages}{4604--4616}.
\newblock \URLprefix \url{https://aclanthology.org/2023.acl-long.253},
  \DOIprefix\doi{10.18653/v1/2023.acl-long.253}.
\bibitem[{Tang et~al.(2023c)Tang, Zhang, Loakman, Lin and
  Guerin}]{tang2023terminology}
\bibinfo{author}{Tang, C.}, \bibinfo{author}{Zhang, H.},
  \bibinfo{author}{Loakman, T.}, \bibinfo{author}{Lin, C.},
  \bibinfo{author}{Guerin, F.}, \bibinfo{year}{2023}c.
\newblock \bibinfo{title}{Terminology-aware medical dialogue generation}, in:
  \bibinfo{booktitle}{ICASSP 2023-2023 IEEE International Conference on
  Acoustics, Speech and Signal Processing (ICASSP)},
  \bibinfo{organization}{IEEE}. pp. \bibinfo{pages}{1--5}.
\bibitem[{Tang et~al.(2022c)Tang, Zhang, Loakman, Lin and
  Guerin}]{tang2022ngep}
\bibinfo{author}{Tang, C.}, \bibinfo{author}{Zhang, Z.},
  \bibinfo{author}{Loakman, T.}, \bibinfo{author}{Lin, C.},
  \bibinfo{author}{Guerin, F.}, \bibinfo{year}{2022}c.
\newblock \bibinfo{title}{{NGEP}: A graph-based event planning framework for
  story generation}, in: \bibinfo{booktitle}{Proceedings of the 2nd Conference
  of the Asia-Pacific Chapter of the Association for Computational Linguistics
  and the 12th International Joint Conference on Natural Language Processing
  (Volume 2: Short Papers)}, \bibinfo{publisher}{Association for Computational
  Linguistics}, \bibinfo{address}{Online only}. pp. \bibinfo{pages}{186--193}.
\newblock \URLprefix \url{https://aclanthology.org/2022.aacl-short.24}.
\bibitem[{Wang et~al.(2021)Wang, Lin, Liu and Wong}]{wang-etal-2021-fast}
\bibinfo{author}{Wang, D.}, \bibinfo{author}{Lin, C.}, \bibinfo{author}{Liu,
  Q.}, \bibinfo{author}{Wong, K.F.}, \bibinfo{year}{2021}.
\newblock \bibinfo{title}{Fast and scalable dialogue state tracking with
  explicit modular decomposition}, in: \bibinfo{booktitle}{Proceedings of the
  2021 Conference of the North American Chapter of the Association for
  Computational Linguistics: Human Language Technologies},
  \bibinfo{address}{Online}. pp. \bibinfo{pages}{289--295}.
\newblock \URLprefix \url{https://aclanthology.org/2021.naacl-main.27},
  \DOIprefix\doi{10.18653/v1/2021.naacl-main.27}.
\bibitem[{Woodsend and Lapata(2010)}]{woodsend-lapata-2010-automatic}
\bibinfo{author}{Woodsend, K.}, \bibinfo{author}{Lapata, M.},
  \bibinfo{year}{2010}.
\newblock \bibinfo{title}{Automatic generation of story highlights}, in:
  \bibinfo{booktitle}{Proceedings of the 48th Annual Meeting of the Association
  for Computational Linguistics}, \bibinfo{address}{Uppsala, Sweden}. pp.
  \bibinfo{pages}{565--574}.
\newblock \URLprefix \url{https://aclanthology.org/P10-1058}.
\bibitem[{Xing et~al.(2020)Xing, Fan and Wan}]{xing-etal-2020-automatic}
\bibinfo{author}{Xing, X.}, \bibinfo{author}{Fan, X.}, \bibinfo{author}{Wan,
  X.}, \bibinfo{year}{2020}.
\newblock \bibinfo{title}{Automatic generation of citation texts in scholarly
  papers: A pilot study}, in: \bibinfo{booktitle}{Proceedings of the 58th
  Annual Meeting of the Association for Computational Linguistics},
  \bibinfo{address}{Online}. pp. \bibinfo{pages}{6181--6190}.
\newblock \URLprefix \url{https://aclanthology.org/2020.acl-main.550},
  \DOIprefix\doi{10.18653/v1/2020.acl-main.550}.
\bibitem[{Xu et~al.(2020)Xu, Patwary, Shoeybi, Puri, Fung, Anandkumar and
  Catanzaro}]{xu-etal-2020-megatron}
\bibinfo{author}{Xu, P.}, \bibinfo{author}{Patwary, M.},
  \bibinfo{author}{Shoeybi, M.}, \bibinfo{author}{Puri, R.},
  \bibinfo{author}{Fung, P.}, \bibinfo{author}{Anandkumar, A.},
  \bibinfo{author}{Catanzaro, B.}, \bibinfo{year}{2020}.
\newblock \bibinfo{title}{{MEGATRON}-{CNTRL}: Controllable story generation
  with external knowledge using large-scale language models}, in:
  \bibinfo{booktitle}{Proceedings of the 2020 Conference on Empirical Methods
  in Natural Language Processing (EMNLP)}, \bibinfo{address}{Online}. pp.
  \bibinfo{pages}{2831--2845}.
\newblock \URLprefix \url{https://aclanthology.org/2020.emnlp-main.226},
  \DOIprefix\doi{10.18653/v1/2020.emnlp-main.226}.
\bibitem[{Yang et~al.(2023a)Yang, Tang and Lin}]{yang2023improving}
\bibinfo{author}{Yang, B.}, \bibinfo{author}{Tang, C.}, \bibinfo{author}{Lin,
  C.}, \bibinfo{year}{2023}a.
\newblock \bibinfo{title}{Improving medical dialogue generation with abstract
  meaning representations}.
\newblock \bibinfo{journal}{arXiv preprint arXiv:2309.10608} .
\bibitem[{Yang et~al.(2023b)Yang, Tang, Zhao, Xiao and Lin}]{yang2023effective}
\bibinfo{author}{Yang, B.}, \bibinfo{author}{Tang, C.}, \bibinfo{author}{Zhao,
  K.}, \bibinfo{author}{Xiao, C.}, \bibinfo{author}{Lin, C.},
  \bibinfo{year}{2023}b.
\newblock \bibinfo{title}{Effective distillation of table-based reasoning
  ability from llms}.
\newblock \bibinfo{journal}{arXiv preprint arXiv:2309.13182} .
\bibitem[{Yao et~al.(2019)Yao, Peng, Weischedel, Knight, Zhao and
  Yan}]{yao2019plan}
\bibinfo{author}{Yao, L.}, \bibinfo{author}{Peng, N.},
  \bibinfo{author}{Weischedel, R.}, \bibinfo{author}{Knight, K.},
  \bibinfo{author}{Zhao, D.}, \bibinfo{author}{Yan, R.}, \bibinfo{year}{2019}.
\newblock \bibinfo{title}{Plan-and-write: Towards better automatic
  storytelling}, in: \bibinfo{booktitle}{AAAI'19}, \bibinfo{publisher}{AAAI
  Press}. pp. \bibinfo{pages}{0--1}.
\newblock \URLprefix \url{https://doi.org/10.1609/aaai.v33i01.33017378},
  \DOIprefix\doi{10.1609/aaai.v33i01.33017378}.
\bibitem[{You et~al.(2020)You, Sun and Iyyer}]{you-etal-2020-hard}
\bibinfo{author}{You, W.}, \bibinfo{author}{Sun, S.}, \bibinfo{author}{Iyyer,
  M.}, \bibinfo{year}{2020}.
\newblock \bibinfo{title}{Hard-coded {G}aussian attention for neural machine
  translation}, in: \bibinfo{booktitle}{Proceedings of the 58th Annual Meeting
  of the Association for Computational Linguistics}, \bibinfo{address}{Online}.
  pp. \bibinfo{pages}{7689--7700}.
\newblock \URLprefix \url{https://aclanthology.org/2020.acl-main.687},
  \DOIprefix\doi{10.18653/v1/2020.acl-main.687}.
\bibitem[{Zhang et~al.(2023)Zhang, Tang, Loakman, Lin and
  Goetze}]{zhanga2023cadge}
\bibinfo{author}{Zhang, H.}, \bibinfo{author}{Tang, C.},
  \bibinfo{author}{Loakman, T.}, \bibinfo{author}{Lin, C.},
  \bibinfo{author}{Goetze, S.}, \bibinfo{year}{2023}.
\newblock \bibinfo{title}{Cadge: Context-aware dialogue generation enhanced
  with graph-structured knowledge aggregation}.
\newblock \bibinfo{journal}{arXiv preprint arXiv:2305.06294} .
\bibitem[{Zhu et~al.(2015)Zhu, Kiros, Zemel, Salakhutdinov, Urtasun, Torralba
  and Fidler}]{zhu2015aligning}
\bibinfo{author}{Zhu, Y.}, \bibinfo{author}{Kiros, R.}, \bibinfo{author}{Zemel,
  R.}, \bibinfo{author}{Salakhutdinov, R.}, \bibinfo{author}{Urtasun, R.},
  \bibinfo{author}{Torralba, A.}, \bibinfo{author}{Fidler, S.},
  \bibinfo{year}{2015}.
\newblock \bibinfo{title}{Aligning books and movies: Towards story-like visual
  explanations by watching movies and reading books}, in:
  \bibinfo{booktitle}{Proceedings of the IEEE international conference on
  computer vision}, pp. \bibinfo{pages}{19--27}.

\end{thebibliography}



\end{document}